%% file: main.tex
\documentclass{article}

\usepackage{xcolor}
\usepackage[preprint]{corl_2026} 

\usepackage{natbib}
\usepackage{graphicx}
\usepackage{caption}
\usepackage{subcaption}
\usepackage{wrapfig}
\usepackage{algorithm}
\usepackage{siunitx}
\usepackage{xspace}

\floatname{algorithm}{Code example}
\usepackage{listings}

\usepackage{fancyhdr}
\usepackage{url}
\usepackage{booktabs}

\definecolor{pyKeyword}{RGB}{0,0,200}     
\definecolor{pyComment}{RGB}{0,128,0}     
\definecolor{pyString}{RGB}{170,55,40}    
\definecolor{pyBuiltin}{RGB}{128,0,128}   
\definecolor{pyClass}{RGB}{0,128,128}     

\lstdefinestyle{python}{
    language=Python,
    basicstyle=\ttfamily\small,
    keywordstyle=\color{pyKeyword}\bfseries,
    commentstyle=\color{pyComment}\itshape,
    stringstyle=\color{pyString},
    showstringspaces=false,
    breaklines=true,
    columns=fullflexible,
    morekeywords=[2]{print, dict, from_dict},
    keywordstyle=[2]\color{pyBuiltin},
    emph={OrcaHand, OrcaJointPosition, OrcaJointPositions,
          MockOrcaHand, OrcaHandTouch, SimOrcaHand,
          OrcaHandRight, OrcaHandRightCubeOrientation, OrcaPandaCubeStacking,
          OrcaHandSink, OrcaHandSimSink,
          MetaQuestPublisher, MediaPipePublisher,
          Retargeter, TargetPose, LeRobotPolicyAdapter},
    emphstyle=\color{pyClass}\bfseries,
}
\input{helpers.tex}

\title{\mytitle}

\input{authors}

\begin{document}
\maketitle

\begingroup
\renewcommand\thefootnote{}
\footnotetext{Code available at: \url{https://github.com/orcahand}}
\endgroup

\input{sections/00_abstract.tex}


\section{Introduction}
\input{sections/01_introduction}

\section{Open Hardware}
\input{sections/02_hardware}

\section{A Platform for Dexterity}
\input{sections/03_software}

\section{Platform Capabilities}
\input{sections/04_features}

\section{Conclusions}
\input{sections/05_conclusions}

\clearpage

\bibliography{references}  

\clearpage
\appendix
\section{The \orca~Stack in Practice}
\input{sections/06_appendix}

\end{document}

%% file: helpers.tex
\newcommand{\orca}{\texttt{orca}\xspace}
\newcommand{\orcacore}{\texttt{orca\_core}\xspace}
\newcommand{\orcasim}{\texttt{orca\_sim}\xspace}
\newcommand{\orcateleop}{\texttt{orca\_teleop}\xspace}
\newcommand{\orcaarm}{\texttt{orca\_arm}}
\newcommand{\lerobot}{\texttt{lerobot}\xspace}

\newcommand{\mytitle}{%
    \orca: A Platform for Open-Source \\ Dexterity Research
}

%% file: authors.tex
\author{
  Francesco Capuano\\
  University of Oxford
  \And
  Maximilian Eberlein \\
  ETH Zurich
  \And 
  Fabrice Bourquin \\
  ETH Zurich
  \And 
  Clemens C. Christoph \\
  Orca Dexterity
}

%% file: sections/00_abstract.tex
\begin{abstract}
    Robotics manipulation research increasingly focuses on two-finger parallel grippers for their effectiveness, affordability, and ease of teleoperation. 
    Grippers are nonetheless limited by their form factor, often requiring bimanual setups even for simple reorientation tasks.
    Anthropomorphic hands are a more natural platform for dexterous robot learning---closer to the human hand, and capable of learning from human video---yet they remain hard to use in learning research: even where open and accessible hand hardware exists, the software for control, simulation, teleoperation, and retargeting is scattered in one-off code bases, and largely disconnected from the robot-learning ecosystem.
    In this work, we introduce the \orca~learning stack, an open-source research stack for dexterity as a first-class robot learning domain.
    Our \orca~stack unifies low-level control, simulation, teleoperation from a range of consumer platforms, and hand retargeting, behind a single interface, and integrates natively with popular robot-learning frameworks such as \lerobot, so dexterous hand researchers can leverage the same data, training, and evaluation pipelines used for non-dexterous robot learning.
    We demonstrate a complete end-to-end workflow, collecting expert demonstrations of an in-hand reorientation task by teleoperation with a consumer-grade VR headset, training an autonomous policy with \lerobot, and evaluating the learned policy in a fully reproducible and observable setup.
    We open-source the entire stack as a shared, reproducible foundation for dexterous-manipulation research.
\end{abstract}

%% file: sections/01_introduction.tex
Robot manipulation research is increasingly built around parallel-jaw grippers~\citep{brohanRT1RoboticsTransformer2023a,brohanRT2VisionLanguageActionModels2023a,kimOpenVLAOpenSourceVisionLanguageAction2024a,chiDiffusionPolicyVisuomotor2024a,zhaoLearningFineGrainedBimanual2023a,shukorSmolVLAVisionLanguageActionModel2025a}.
Grippers are indeed reliable, mechanically simple, effective end effectors, and have been widely used by the research community to perform tasks such as moving objects in a scene~\citep{liuLIBEROBenchmarkingKnowledge2023a}, pushing them with a stick-like end-effector~\citep{florenceImplicitBehavioralCloning2021}, or perform more complex, fine-grained manipulation tasks~\citep{zhaoLearningFineGrainedBimanual2023a,black$p_0$VisionLanguageActionFlow2026}. 
Yet, human-level dexterity remains elusive, as it likely demands the many contacts and degrees of freedom of hands, making anthropomorphic hand designs a more promising path toward highly-dexterous, autonomous robots.

Grippers are easy to teleoperate: \textit{leader-follower schemes}~\citep{zhaoLearningFineGrainedBimanual2023a,wuGELLOGeneralLowCost2024a,cadeneLeRobotOpenSourceLibrary2026} let an operator drive a robot arm directly, often even via a simplified, low-cost replica of the follower~\citep{wuGELLOGeneralLowCost2024a}.
In turn, teleoperation pipelines have been pivotal in producing the large demonstration datasets~\citep{collaborationOpenXEmbodimentRobotic2025, khazatskyDROIDLargeScaleInTheWild2024, cadeneLeRobotOpenSourceLibrary2026} used to train large-scale \emph{foundation models for arms with grippers}~\citep{brohanRT1RoboticsTransformer2023a,brohanRT2VisionLanguageActionModels2023a,kimOpenVLAOpenSourceVisionLanguageAction2024a,black$p_0$VisionLanguageActionFlow2026,shukorSmolVLAVisionLanguageActionModel2025a}.
On a more practical note, grippers are also cheap to maintain: their mechanical simplicity localizes failures to a few visible, typically inexpensive wear parts---pads, fingertips, linkages, or commodity actuators---whereas dexterous hands pack many actuators, transmissions, sensors, and contact surfaces into a compact volume, making wear harder to locate and repair.
In turn, lower operating costs mitigate the friction towards adopting grippers over anthropomorphic hands in research~\citep{vedderCaseHumanHands}.
Still, dexterous hands can in principle unlock manipulation capabilities that grippers simply cannot reach.
Indeed, multi-fingered hands can perform in-hand reorientation tasks and other contact-rich skills~\citep{chenSystemGeneralInHand2021,kediaSimToolRealObjectCentricPolicy2026a} that grippers can only achieve through rather elaborate control strategies (e.g.,~\citet{antonovaReinforcementLearningPivoting2017a}), if at all.
As they share human morphology, hands can also directly use tools, interfaces, and environments built for humans with minor task-specific adaptation, and support compliant, intuitive contact in shared workspaces.
While in principle coordinating several gripper arms could substitute for one hand's dexterity, teleoperating more than two arms remains impractical, blocking wider adoption.

\begin{figure}
    \centering
    \includegraphics[width=0.99\linewidth]{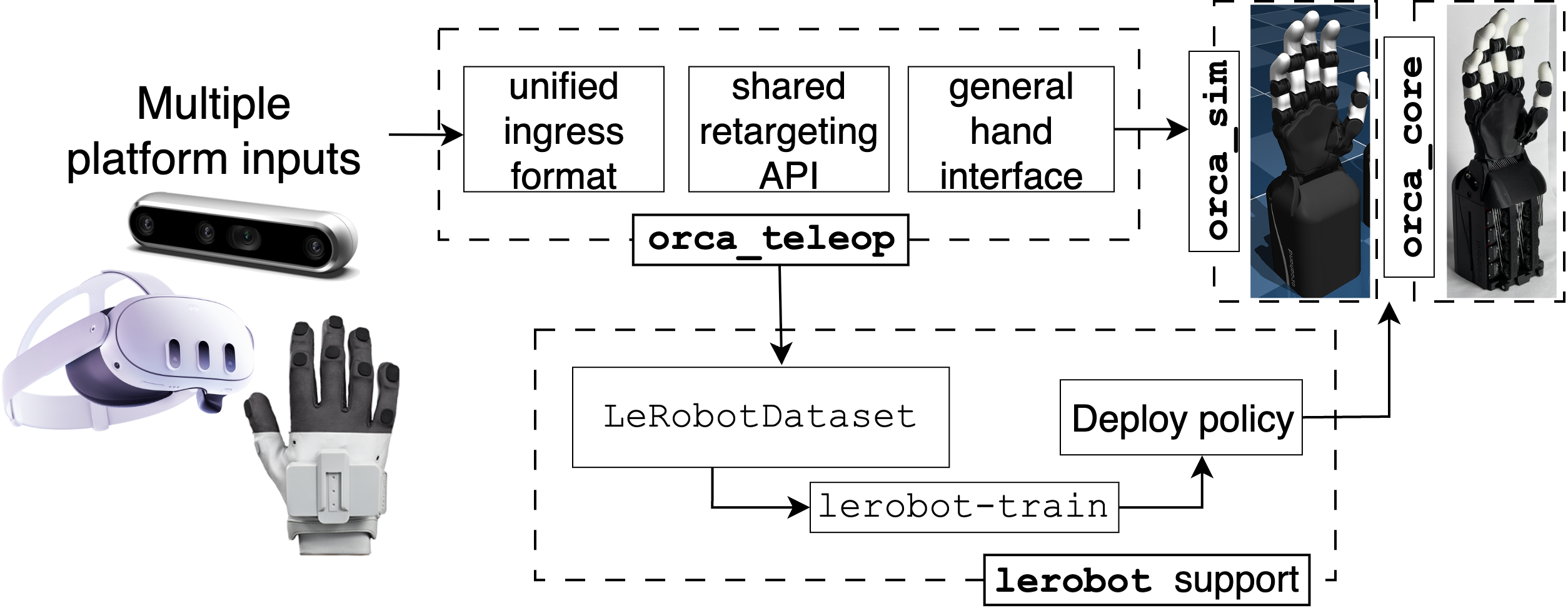}
    \caption{\texttt{orca} is an open-source, end-to-end stack for dexterous robot learning. We release a single software platform unifying control, simulation, teleoperation, and human-to-robot retargeting, integrating natively with popular robot learning tools such as \texttt{lerobot}.}
    \label{fig:fig1}
\end{figure}

Further, hands minimize the embodiment gap between robots and human demonstrators.
A human-like hand can in principle be trained directly on recordings of people performing the same task~\citep{kareerEgoMimicScalingImitation2024}, without the necessity of building and maintaining teleoperation setups.
This promises demonstration data at a new scale, incorporating in-the-wild human video as weak supervision~\citep{kareerEgoMimicScalingImitation2024, zhengEgoscaleScalingDexterous2026, yangEgoVLALearningVisionLanguageAction2025} or low-cost capture setups made viable by the kinematic similarity between human and robot hands~\citep{wangDexCapScalablePortable2024}.

Despite all these advantages, dexterous hands are still uncommon in robot-learning research due to their cost and integration challenges: with no unified, accessible, open-source software stack for robot hands, progress in research is fundamentally hindered by the lack of an accessible, open platform for dexterity.
Researchers typically assemble bespoke stacks from isolated off-the-shelf components, including hand-specific SDKs, one-off retargeting scripts, and custom teleoperation pipelines tied to a particular input device and robot. 
Because these components are seldom designed to interoperate, reproducing prior work often requires substantial integration effort, hindering progress.
This work introduces \orca, an open-source, end-to-end software stack for dexterous-manipulation research, aiming at addressing precisely these limitations.
The stack is built around the Orcahand~\citep{christophORCAOpenSourceReliable2025}, a tendon-driven, fully 3D-printable hand costing ca. \$3k---a fraction of higher end alternatives such as the SHARPA (ca. \$50k) or Shadow Dexterous (ca. \$60k) hands.
In this, we contribute a software stack turning a highly accessible dexterous hand into a first-class robot-learning platform, aiming at democratizing dexterity research.
%
In summary, our contributions are:
\begin{itemize}
    \item \textbf{The \orca~learning stack}, a fully open-source research stack that turns dexterous hand into first-class robot-learning platforms. We unify control, teleoperation, human-to-robot retargeting, and simulation, with utilities to train expressive in-hand policies, as well as supporting the \orca~hand's integration with manipulators such as the OpenArm~\citep{openarm} and Franka Panda~\citep{frankaemika_panda} arms.
    \item \textbf{Porting dexterity research to \lerobot}~\citep{cadeneLeRobotOpenSourceLibrary2026}, enabling policies for dexterous hands to be trained and evaluated with the same tooling and abstractions the robot learning community already uses for other highly accessible, low-cost robot platforms.
\end{itemize}

\clearpage

%% file: sections/02_hardware.tex
\begin{figure}[t]
    \centering
    \includegraphics[width=\linewidth]{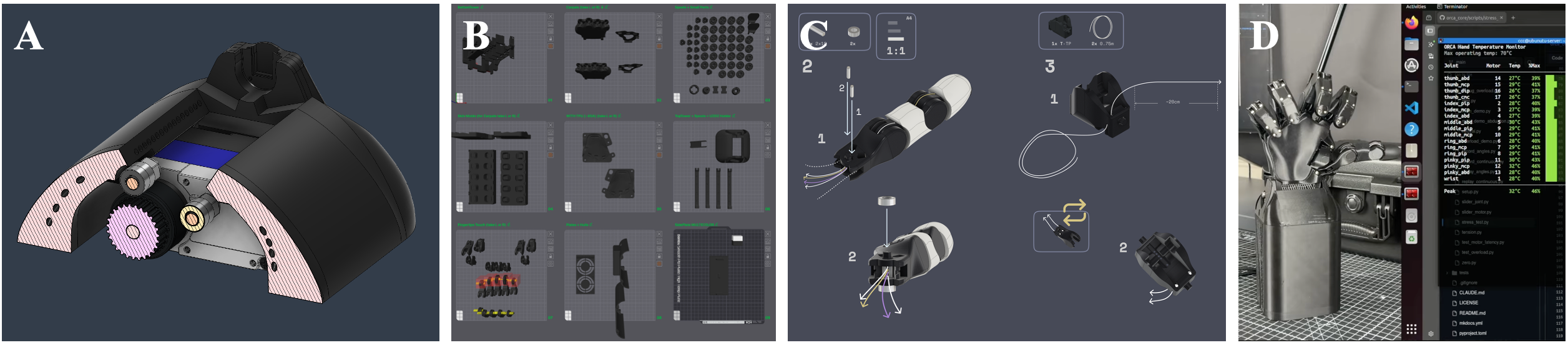}
    \caption{The Orcahand hardware is fully open, with releases including (A) Fully adjustable Autodesk Fusion CAD file (B) Bambu Lab 3D print files with fine-tuned parameters. (C) Step-by-step assembly instructions (D) Getting-started tutorials and diagnostic tools. Open, highly accessible hardware is a cornerstone of the \orca~stack for dexterity, empowering researchers despite their resources.}
    \label{fig:hardware_overview}
\end{figure}

\begin{wrapfigure}{r}{0.38\linewidth}
    \centering
    \includegraphics[width=0.9\linewidth]{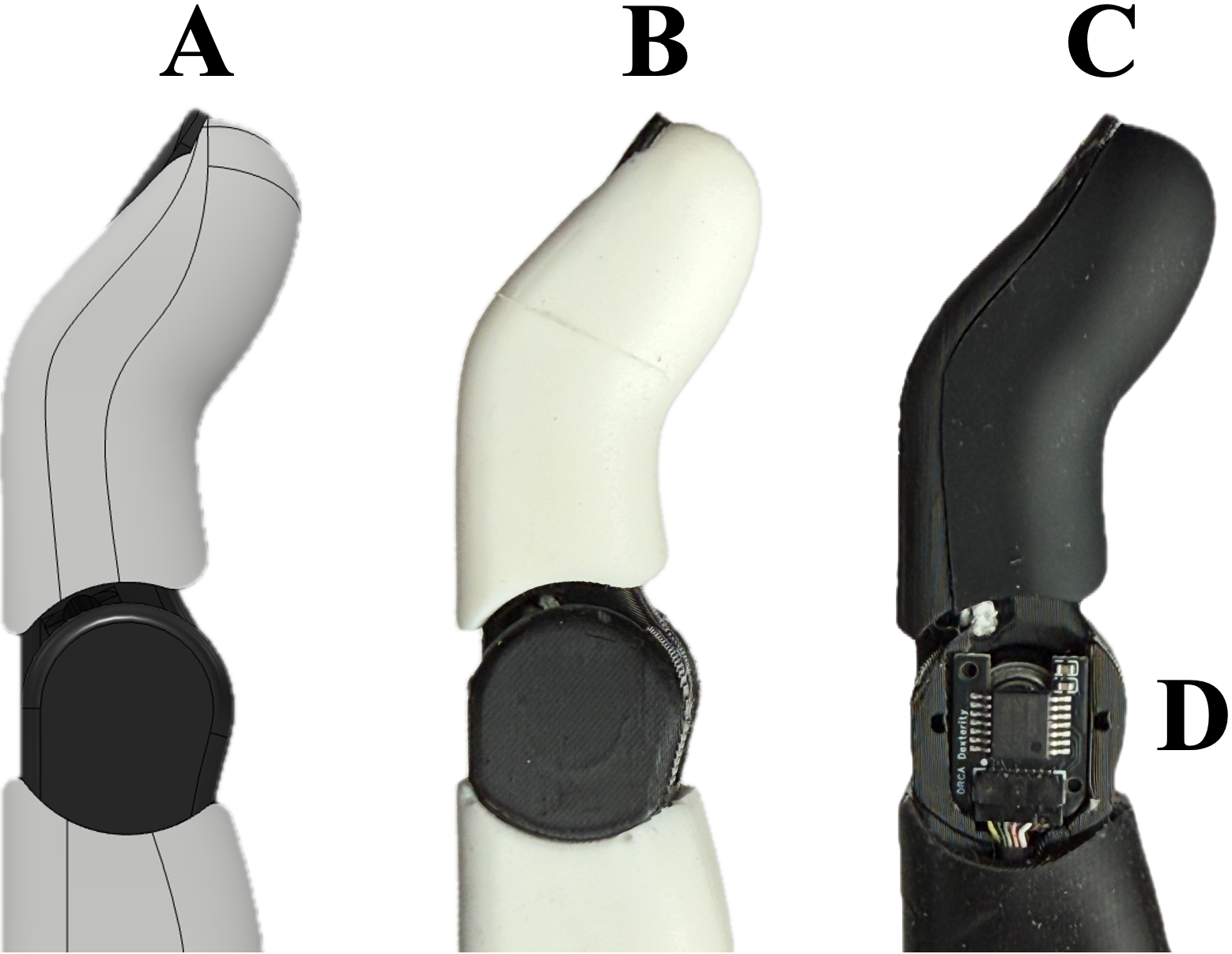}
    \caption{Fingertip variants of the Orcahand all share the same geometries. (A) CAD/URDF model. (B) 3D-printed fingertip with soft silicone skin. (C) Fingertip with integrated tactile sensor, and silicone skin. (D) Improved proprioception joint.}
    \label{fig:fingertip_variants}
\end{wrapfigure}

Our robot learning stack is built around the Orcahand, an openly available (Figure~\ref{fig:hardware_overview}) robot hand, whose design is guided by five main principles: (1) accessibility, (2) affordability, (3) antropomorphism, (4) ease of assembly, and (5) repairability. 
Every part of the Orcahand can be either 3D-printed or acquired off-the-shelf, making the whole hand affordable and research-friendly, compared to other dexterous robotic hands.
The open-source hardware design allows researchers to easily replace parts of the hand themselves in case of damage or wear, rather than having to send the hand back for repairs.
Specifically, the Orcahand has 17 degrees of freedom (DoF, 1 for wrist, 16 for fingers): the wrist joint is actuated by a belt-driven mechanism, while each finger joints is fully actuated by an antagonistic tendon-pair~\citep{christophORCAOpenSourceReliable2025}.
The range of motion of each joint as well as the finger morphologies are designed to be similar to that of a human hand, facilitating learning from human videos, teleoperation and natural interaction with objects specifically built for humans.

The Orcahand is available in different variants, all sharing the same fundamental mechanical design and geometry, but differing in whether (1) they support tactile sensing, (2) advanced proprioception, and (3) their actuation capabilities (Figure~\ref{fig:fingertip_variants}).
First, the fingertips can be equipped with custom-manufactured Hall-effect 6D force/torque sensors (Figure~\ref{fig:fingertip_variants}, C), providing 51 taxels-per-digit on the thumb and pinky, and 83 taxels-per-digit on the remaining fingers.
Second, the hand can optionally be built with miniaturized encoders integrated in each joint (Figure~\ref{fig:fingertip_variants}, D) enabling improved proprioception and closed-loop control.
Third, the Orcahand platform supports two different motor models: the research-standard Dynamixel XC330-T288 and XC430-T240BB, and the more cost-effective Feetech HL-3915 and HL-3930.
All variants are supported by the same \orca~software stack, further underscoring the adaptability and modularity of our contribution.

%% file: sections/03_software.tex
Currently, the tooling needed to carry out dexterous-hand research is fragmented across single-purpose codebases.
In the typical research workflow, one typically has to stitch together vendor SDKs for the low-level control of the motors, a variety of libraries for hand tracking, forward and inverse kinematics, and retargeting, adapting standalone teleoperation pipelines for one's own goals, and often needing to build one's own simulation assets.
In turn, this research practice directly hinders reproducibility, and increases fragmentation in the field, as components that are written in isolation expose incompatible interfaces, rarely sharing a common data format.
In a recent open-source effort,~\citet{cadeneLeRobotOpenSourceLibrary2026} standardize much of this plumbing, but crucially target simpler platforms such as robot arms with grippers end effectors, rather than dexterous hands, which remain currently unsupported in any open source library focused on affordable robotics.

\begin{wrapfigure}{r}{0.3\linewidth}
    \centering
    \includegraphics[width=\linewidth]{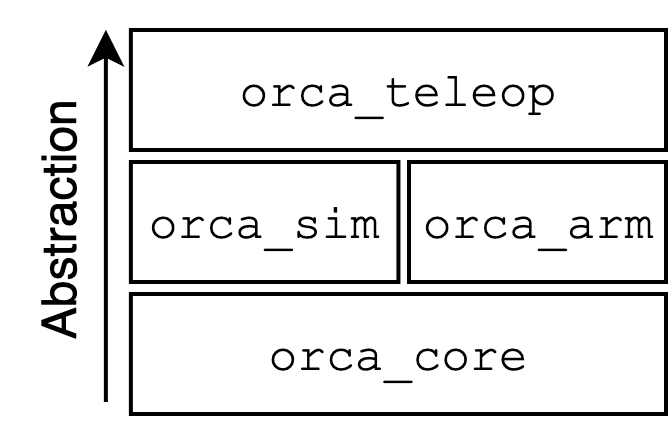}
    \caption{The \orca software stack is organized with increasing levels of abstraction.}
    \label{fig:orca_stack}
\end{wrapfigure}
We address this gap with our \orca~software stack (Figure~\ref{fig:orca_stack}), a single, open research asset for dexterity.
The stack is divided into four interoperating packages that share \emph{one data format} and \emph{one control abstraction}: \orcacore~defines primitives for hardware control, \orcasim~implements simulation, \orcaarm~provides full-arm integration, while \orcateleop~allows for teleoperation and retargeting.
The same policy interface, observation and action spaces are used in simulation and on hardware, so code written against one transfers to the other without modification.
Every package in our stack is openly available and released under the permissive MIT license, and the stack natively integrates with the broader robot-learning ecosystem through \lerobot~\citep{cadeneLeRobotOpenSourceLibrary2026}.
We deliberately split the stack into four narrowly scoped subpackages rather than a monolith package in an effort to empower the community to iterate fast, and efficiently contribute back to our open-source stack.

\begin{figure}
    \centering
    \includegraphics[width=0.75\linewidth]{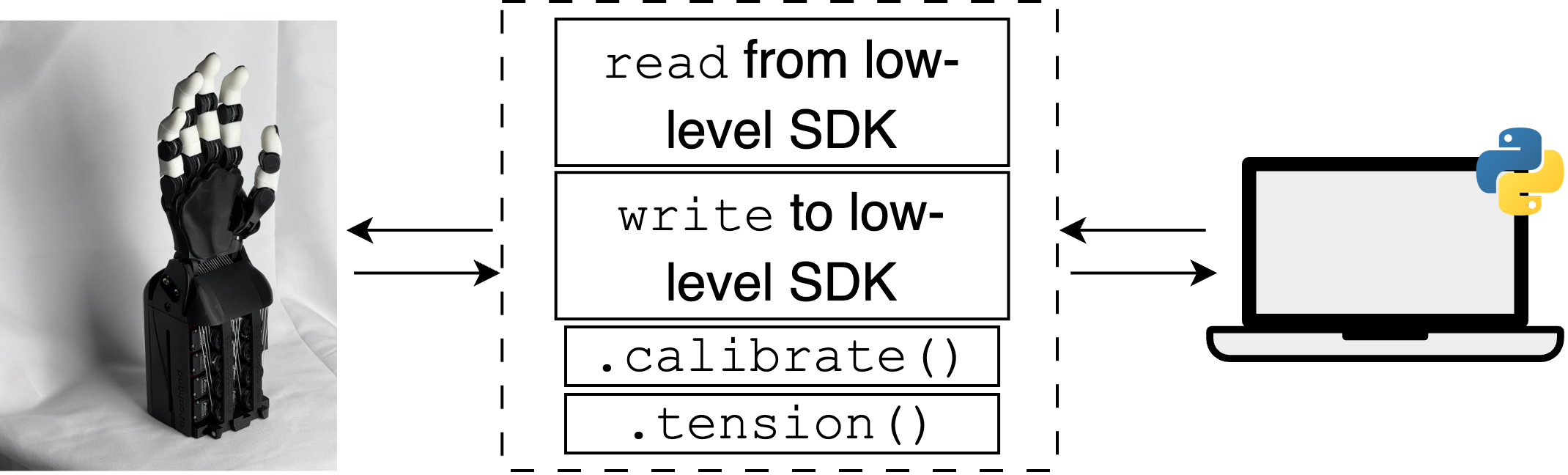}
    \caption{The core package of the \orca stack enables high level access to the internals of both real-world and simulated hands, through a shared, light-weight API.}
    \label{fig:orcacore}
\end{figure}

\paragraph{A shared interface for real and sim hands.}
Our stack is built on a shared, high-level interface to both physical and simulated hands, \orcacore~(Figure~\ref{fig:orcacore}).
This core package exposes position and current control at the joint level following a strongly typed design pattern built around custom types asserted throughout.
In this, \orcacore~effectively abstracts actuation, calibration and tensioning from control, providing a single, joint-space API to build upon.
On top of the raw actuator readings, \orcacore~also reconstructs full joint state, including the sensed joints and potential tactile readings for hands mounting tactile sensors, serving a target hand observations at a fixed control rate suitable for closed-loop policies.
Thus, such lightweight package encapsulates calibration, safety limits, and state estimation, providing the foundation to build higher level hand clients, whether real or simulated.

\paragraph{Simulation support.}

\begin{wrapfigure}[11]{l}{0.25\linewidth}
    \vspace{-20pt}
    \centering
    \includegraphics[width=\linewidth]{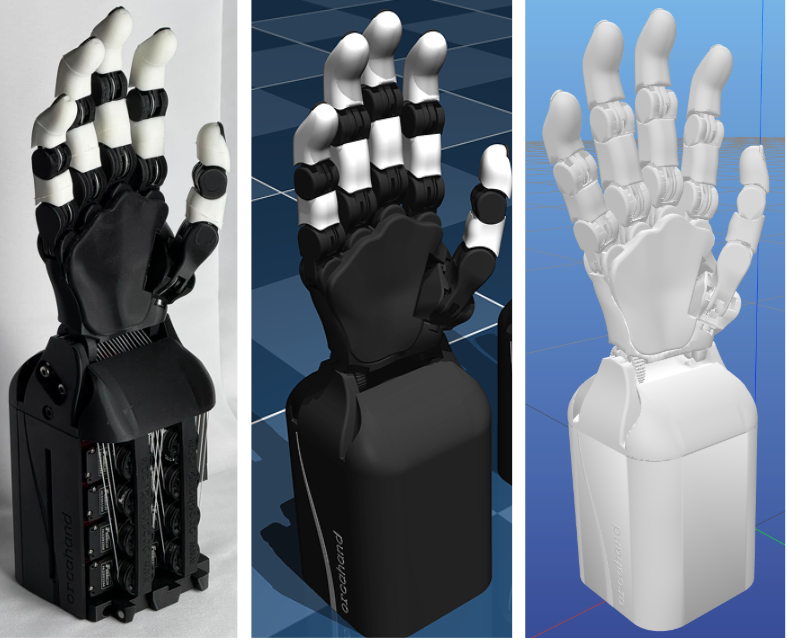}
    \caption{Simulated versions of the real-world hand all share the same description files and low-level API.}
    \label{fig:orcasim}
\end{wrapfigure}

\begin{figure}
    \centering
    \begin{subfigure}[t]{0.32\linewidth}
        \centering
        \includegraphics[height=3.6cm]{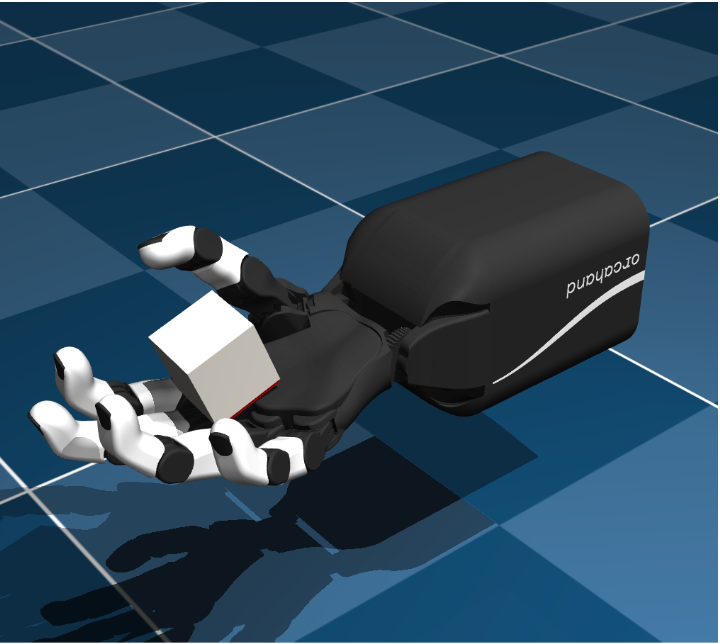}
        \caption{In-hand manipulation task.}
        \label{fig:hand_cube}
    \end{subfigure}
    \hfill
    \begin{subfigure}[t]{0.32\linewidth}
        \centering
        \includegraphics[height=3.6cm]{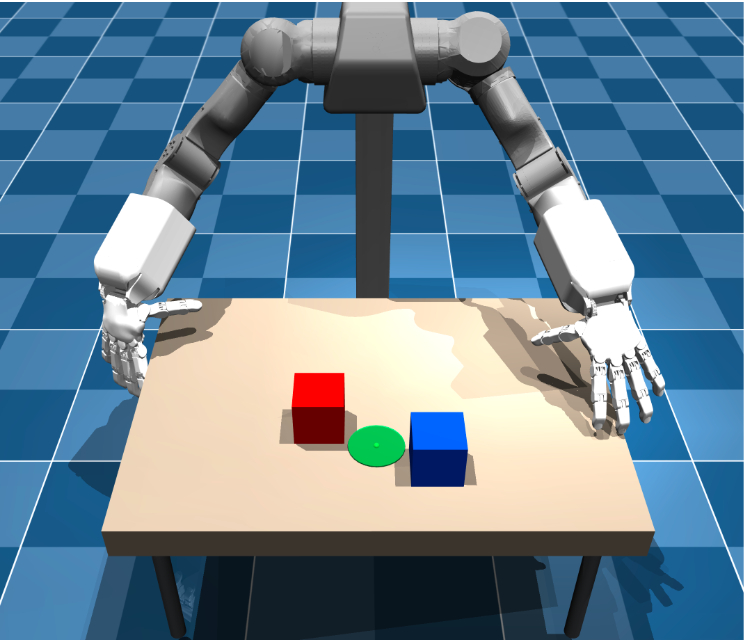}
        \caption{Bimanual OpenArm.}
        \label{fig:orcaarm}
    \end{subfigure}
    \hfill
    \begin{subfigure}[t]{0.32\linewidth}
        \centering
        \includegraphics[height=3.6cm]{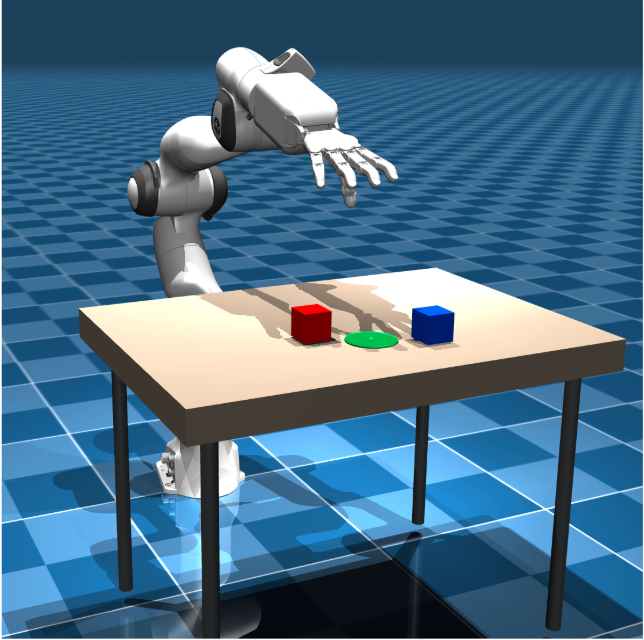}
        \caption{Single-arm Franka Emika Panda.}
        \label{fig:orcapanda}
    \end{subfigure}
    \caption{Simulation of single hand, and full-arm platforms. From left to right: an in-hand manipulation task in \orcasim, a bimanual OpenArm setup with fully dexterous fingers, and a single-arm Franka Emika Panda setup, both available through \orcaarm.}
    \label{fig:orca_platforms}
\end{figure}

\orcasim~extends the real-world hand interface provided in \orcacore~to URDF and MJCF-based simulation.
Starting from the same interfaces, we develop a simulated hand from the same description files (Figure~\ref{fig:orcasim}), and release simulation environments as MuJoCo~\citep{todorovMuJoCoPhysicsEngine2012} models of the hand, together with task-specific environments (Figure~\ref{fig:hand_cube}), allowing users to rapidly and effectively prototype without requiring hardware access.
We also release URDF-based simulation assets, which are integrated with simulation environments such as ManiSkill~\citep{muManiSkillGeneralizableManipulation2021}, which (1) enables GPU-parallel simulation for large-scale training and (2) provides a comprehensive suite of already available manipulation tasks.
Within \orcaarm, we also support complete manipulation platforms for both single-arm, such as the Franka Emika Panda (Figure~\ref{fig:orcapanda}), and bimanual setups, such as OpenArm (Figure~\ref{fig:orcaarm}), enabling integration with task-specific environments (Figure~\ref{fig:orca_platforms}).

\paragraph{Streamlined teleoperation.}

\begin{figure}
    \centering
    \includegraphics[width=\linewidth]{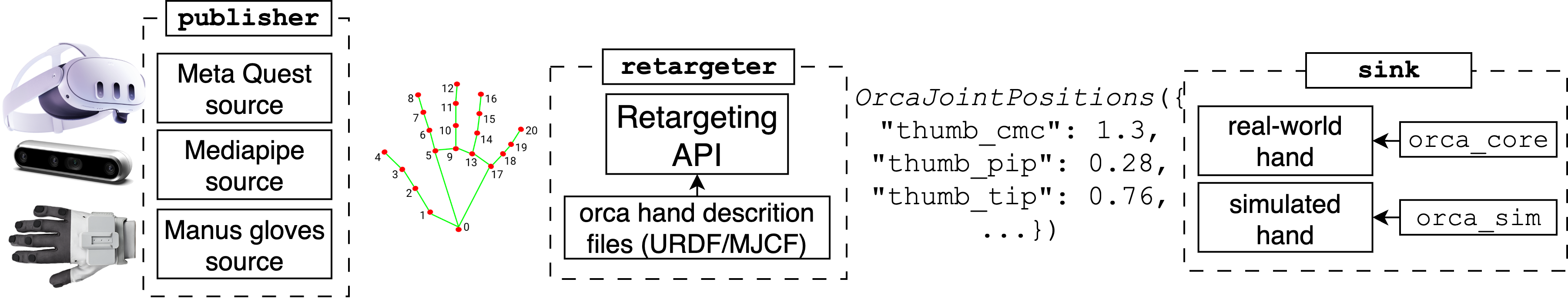}
    \caption{\orcateleop streamlines~teleoperation from consumer grade devices including cameras, VR headsets and mocap gloves, building a shared pipeline for both real and simulated hands.}
    \label{fig:orcateleop}
\end{figure}

Human-level dexterity demands more capable autonomous policies.
While effective, reinforcement learning for robotics~\citep{kober2013reinforcement, antonovaReinforcementLearningPivoting2017a, akkaya2019solving} still struggles in achieving this across tasks, and a broad class of behaviors remains most readily specified via expert human demonstrations~\citep{zhaoLearningFineGrainedBimanual2023a,chiDiffusionPolicyVisuomotor2024a,capuano2025robot, cadeneLeRobotOpenSourceLibrary2026}, which presumes access to a teleoperation interface.
Unfortunately, unlike gripper end-effectors---whose kinematics admit a near-isomorphic leader replica~\citep{zhaoLearningFineGrainedBimanual2023a,wuGELLOGeneralLowCost2024a,cadeneLeRobotOpenSourceLibrary2026}---anthropomorphic hands cannot be as easily teleoperated with actuated replicas.
In turn, teleoperating robot hands requires a pipeline ingesting heterogeneous input streams, and retargeting human motion onto the robot, all under a near real-time constraint to guarantee proper feedback to the robot operator.
\orcateleop~(Figure~\ref{fig:orcateleop}) satisfies these requirements by building directly on the \orcacore~control interface and \orcasim~environments, so that one teleoperation stack can actuate both the physical and the simulated hand, interchangeably.

\orcateleop~runs at 40-50\si{fps}, and supports a range of commodity acquisition devices, including highly accessible camera-based hand tracking via MediaPipe~\citep{lugaresiMediaPipeFrameworkBuilding}, the Meta Quest 3 VR headset~\citep{meta_quest3}, and Manus mocap gloves~\citep{manus_metagloves_pro}.
These disparate input modalities all share the same teleoperation pipeline through a composable abstraction of teleop sources, retargeting operators, and landmarks consumers.
Independently of the input modality, publishers all output the captured hand landmarks in the same convention (Figure~\ref{fig:orcateleop}).
Then, hand landmarks are retargeted onto the hand leveraging the kinematics description provided via URDF assets. 
First, we normalize and scale the human hand pose to guarantee both scale and rotation invariance. 
Then, an optimization routine aims at finding the robot's joint configuration whose forward kinematics (FK) best matches the observed full-hand geometry, expressed with a representation based on key vectors computed from (1) palm pose, (2) fingertip positions, (3) intermediate finger joints, (4) fingertip directions, and (5) augmented with pinch relationships.
The solver leverages kinematics information including joint limits and analytical Jacobians from the FK tree, providing useful for real-time retargeting.
In order to mitigate sensitivity to noisy landmarks and human-robot morphology mismatch, the retargeter employs a Huber loss on the key vectors mismatch, so that the optimization objective is quadratic for small geometric errors, and linear for larger ones, making landmark outliers or morphology mismatch less dominant~\citep{wuji2026retargeting}.
In order to control the full manipulation setups provided in \orcaarm, we also support for wrist poses' matching, reconstructing (or directly capturing) targets (alongside) from the hand landmarks, and matching them with the hand carpals' pose via inverse kinematics.

\orcateleop integrates natively with \lerobot~\citep{cadeneLeRobotOpenSourceLibrary2026}, and allows to both (1) collect expert demonstrations and (2) train autonomous policies based on behavioral cloning.
In practice, demonstrations collected through \orcateleop~are recorded in the \texttt{LeRobotDataset} format, so existing policies supporeted by the library, such as ACT~\citep{zhaoLearningFineGrainedBimanual2023a}, Diffusion Policy~\citep{chiDiffusionPolicyVisuomotor2024a} or \( \pi_0 \)~\citep{black$p_0$VisionLanguageActionFlow2026}, can be trained and deployed with minimal code changes.
In this, we aim at placing dexterous manipulation on the same data, training, and evaluation pipeline that is already used for grippers, effectively lowering the barrier to dexterity research.

%% file: sections/04_features.tex
We substantiate our claims regarding \orca's potential for the research community showcasing an end-to-end dexterous-manipulation workflow: a few lines of code drive the physical hand, its simulated counterpart, and the teleoperation pipeline behind one interface, and the demonstrations collected are used to train a behavioral cloning policy through \lerobot.

\begin{algorithm}
\caption{A minimal \orcacore~usage example.}    
\label{alg:orca-core}
\rule{\linewidth}{1pt}
\begin{lstlisting}
from orca_core import OrcaHand, OrcaJointPosition
hand = OrcaHand()  # <-- auto-detecs connection port

hand.calibrate()  # <-- triggers calibration
hand.tension()  # <-- triggers tendon tensioning

joint_config = {"thumb_cmc": 1.3, "thumb_pip": 0.28, "thumb_tip": 0.76, ...}
hand.set_joint_position(
    OrcaJointPosition.from_dict(**joint_config) # <-- frozen payload
)
\end{lstlisting}
\rule{\linewidth}{1pt}
\end{algorithm}

\begin{algorithm}
\caption{A minimal \orcasim~usage example.}
\label{alg:orca-sim}
\rule{\linewidth}{1pt}
\begin{lstlisting}
from orca_sim import OrcaHandRightCubeOrientation, OrcaPandaCubeStacking

env = OrcaHandRightCubeOrientation()
# gym-like API
obs, info = env.reset()
for _ in range(10):
    action = env.action_space.sample()  # <-- 17-d action vector
    obs, reward, terminated, truncated, info = env.step(action)

    if terminated or trunacted:
        obs, info = env.reset()

env.close()

\end{lstlisting}
\rule{\linewidth}{1pt}
\end{algorithm}

\paragraph{A minimal, shared interface to hands} 
Interfacing a real-world hand with \orca~requires only few lines of Python code (Code example~\ref{alg:orca-core}).
Hands and low-level motor clients are auto-detected scanning over connection ports, and full ROM calibration and tendon tensioning can also be execute with simple one-liners.
The strongly typed hand-interface, \texttt{OrcaJointPosition}, allows assertions on the payloads coming in and out the considered hand, proving useful.
Simulating hands, as well as full manipulation setups, does also take little code (Code example~\ref{alg:orca-sim}).
Users can directly leverage one of the released simulation environments in \orcasim, or adapt the released cross-platform assets (MuJoCo~\citep{todorovMuJoCoPhysicsEngine2012}, ManiSkill~\citep{muManiSkillGeneralizableManipulation2021}, bare URDFs and MJCFs) to build their own.
Simulation environments share the same low-level controls as real-world hands, allowing for fast prototyping without requiring actual hardware, so that third-party consumers can interchangeably use simulated and real-world hands with limited switching costs (Code example~\ref{alg:orca-teleop}).
Teleoperation also reuses the same interface, and supports modular input sources through a common ingress API (Code example~\ref{alg:orca-teleop}).
Hand-tracking frames are retargeted to joint configurations (and optionally solved through IK for full manipulation setups), then dispatched over gRPC to either a real or simulated sink, as \orcacore's shared low-level interface makes the two targets interchangeable from the operator's perspective.

\begin{algorithm}
\caption{A minimal \orcateleop~usage example.}
\label{alg:orca-teleop}
\rule{\linewidth}{1pt}
\begin{lstlisting}
from orca_teleop import pipeline, OrcaHandSink
from orca_teleop.ingress.metaquest import MetaQuestPublisher  # or MediaPipePublisher

# Meta Quest hand tracking -> paced gRPC stream of HandFrames
network_config = {"server_address": "localhost:50051", "quest_host": "localhost"}
publisher = MetaQuestPublisher(**network_config)
publisher.run()

# ingress (gRPC) -> retargeter -> hand landmarks output
sink = OrcaHandSink()  # <-- real-world hand 
pipeline.run(sink=sink, port=50051)

# Alternatively, use a simulated hand sink listening on the same port
sim_sink = OrcaHandSimSink()
pipeline.run(sink=sim_sink, port=50051)

\end{lstlisting}
\rule{\linewidth}{1pt}
\end{algorithm}

\paragraph{Hardware durability.}
\begin{wrapfigure}[11]{r}{0.4\textwidth}
    \vspace{-10pt}
    \centering
    \begin{subfigure}{0.45\linewidth}
        \centering
        \includegraphics[width=\linewidth]{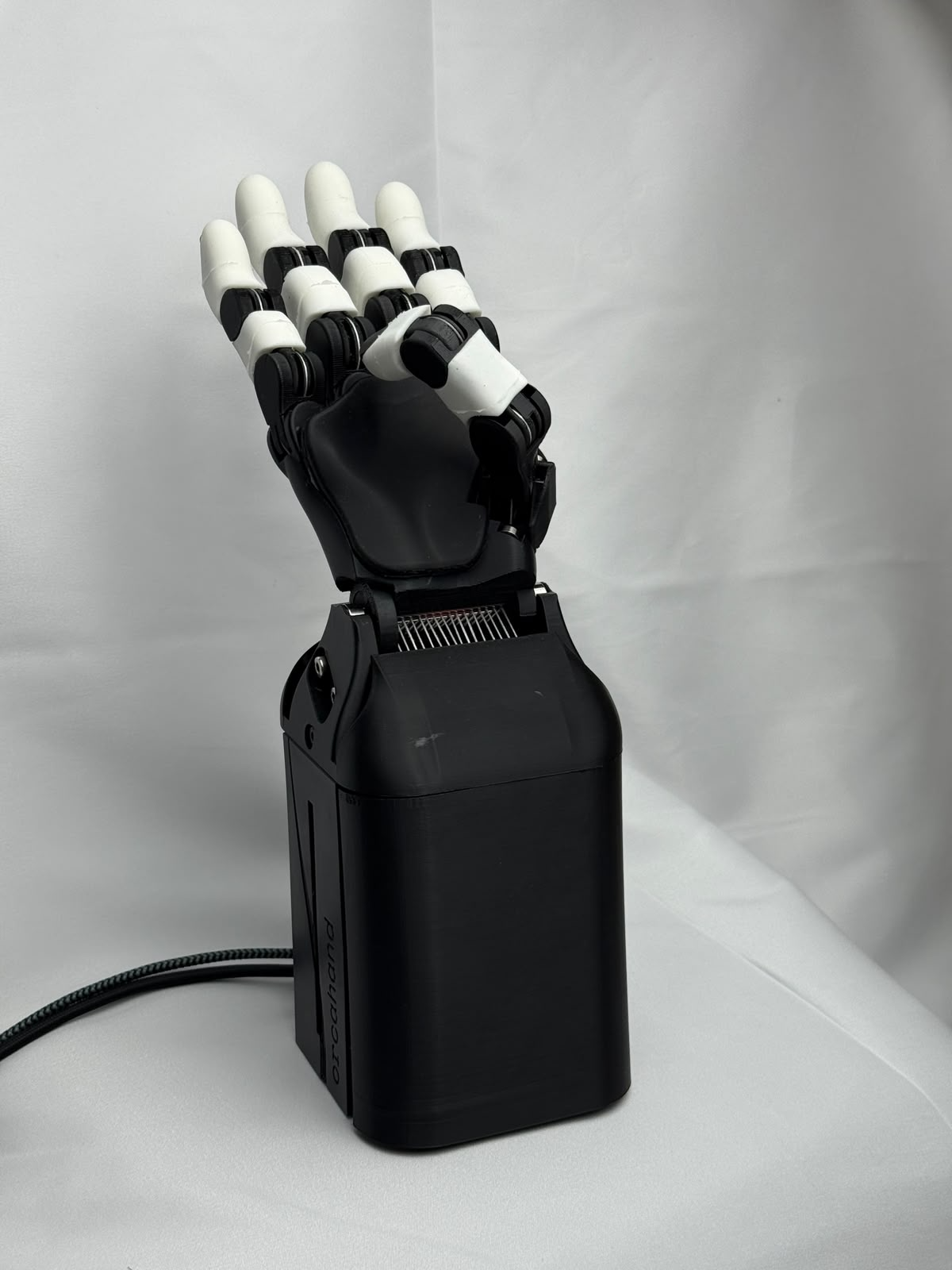}
        \caption{Extended hand pose.}
        \label{fig:hand-open}
    \end{subfigure}
    \begin{subfigure}{0.45\linewidth}
        \centering
        \includegraphics[width=\linewidth]{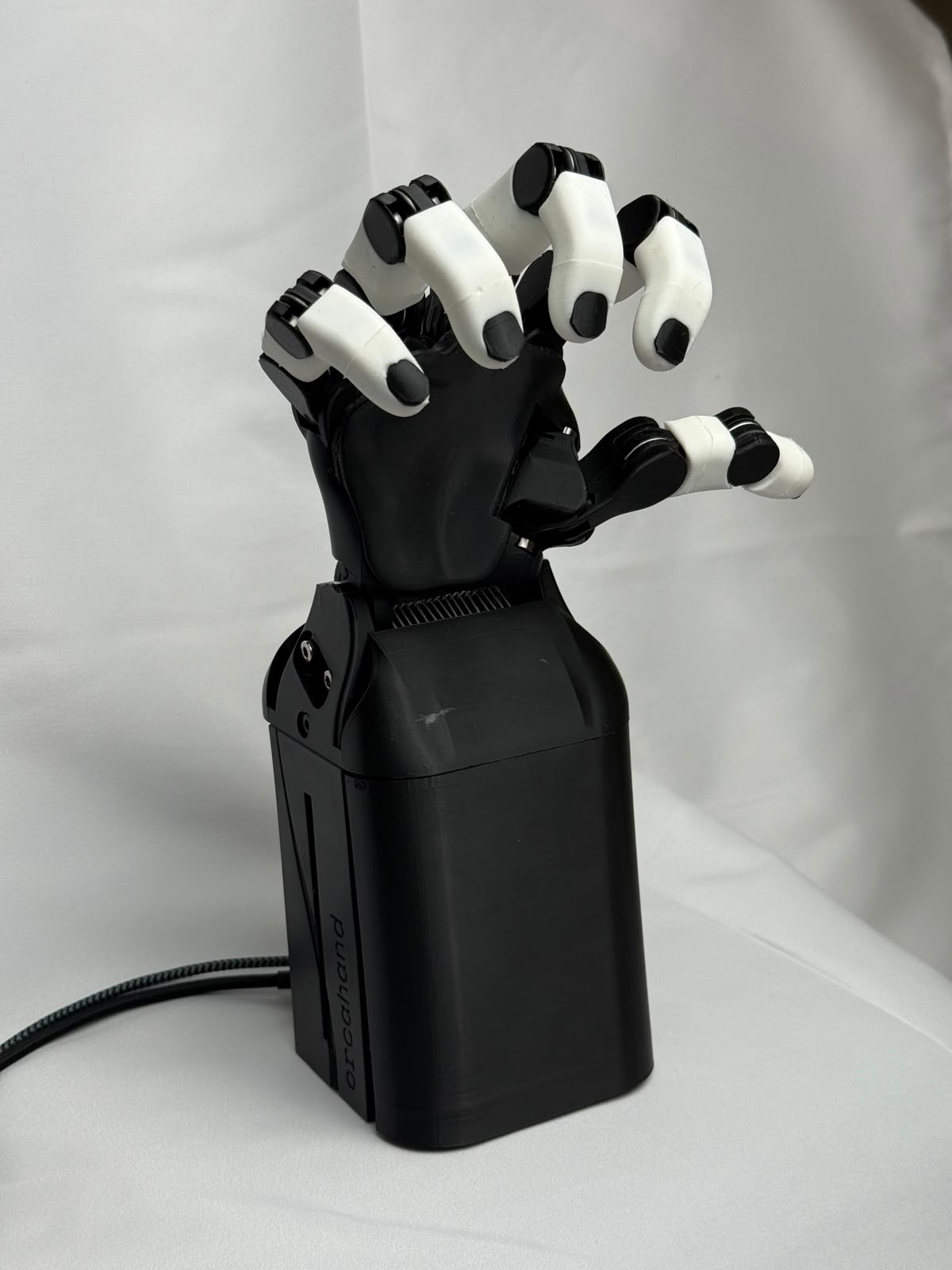}
        \caption{Flexed hand pose.}
        \label{fig:hand-closed}
    \end{subfigure}
\end{wrapfigure}

We characterize the durability of the hand with a reproducible, cyclic stress test. 
The hand is driven repeatedly between an extended (\emph{open}, Figure~\ref{fig:hand-open}) and a flexed (\emph{closed}, Figure~\ref{fig:hand-closed}) configuration, spanning the full joint range of motion across all 17 degrees of freedom, briefly dwelling at each endpoint.
Throughout the run, per-motor temperature is polled at \SI{2}{\hertz} and the test aborts whenever any motor exceeds \SI{70}{\celsius}---a conservative bound below the operating limit of both the Dynamixel XC330/430 and Feetech STS3215 actuators used for the hand.
Over a continuous \SI{5}\hour~session (ca. 3500 open-close cycles), we report no thermal cutoff and no actuator or transmission failure, supporting the durability of the platform under sustained actuation loading.

\begin{wrapfigure}[8]{l}{0.2\textwidth}
    \centering
    \vspace{-2em}
    \includegraphics[width=0.8\linewidth]{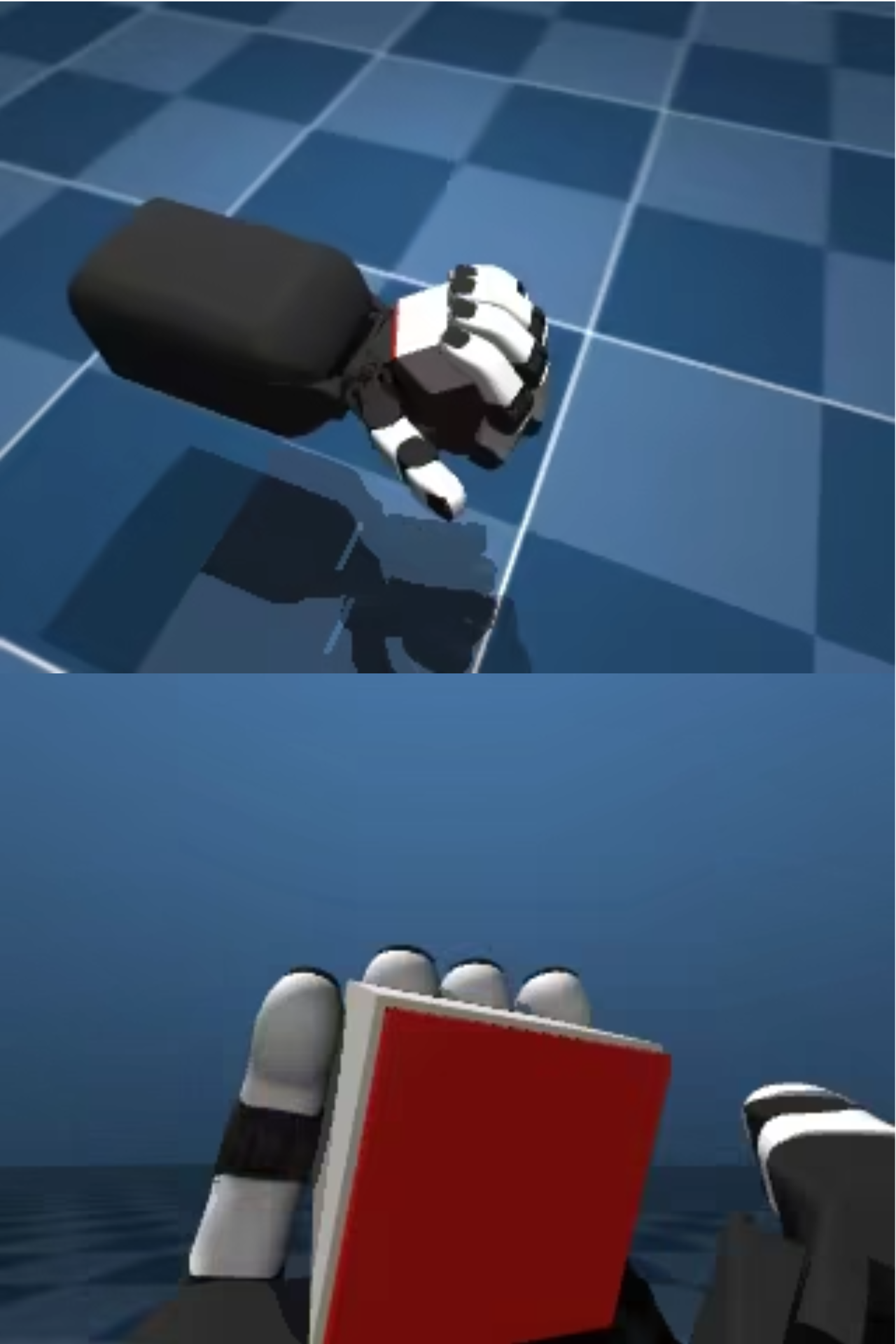}
    \caption{Top and wrist camera views (policy inputs).}
    \label{fig:wrist-and-top}
\end{wrapfigure}
\paragraph{End-to-end teleoperation and imitation learning.}

\begin{figure}[t]
    \centering
    \includegraphics[width=0.99\linewidth]{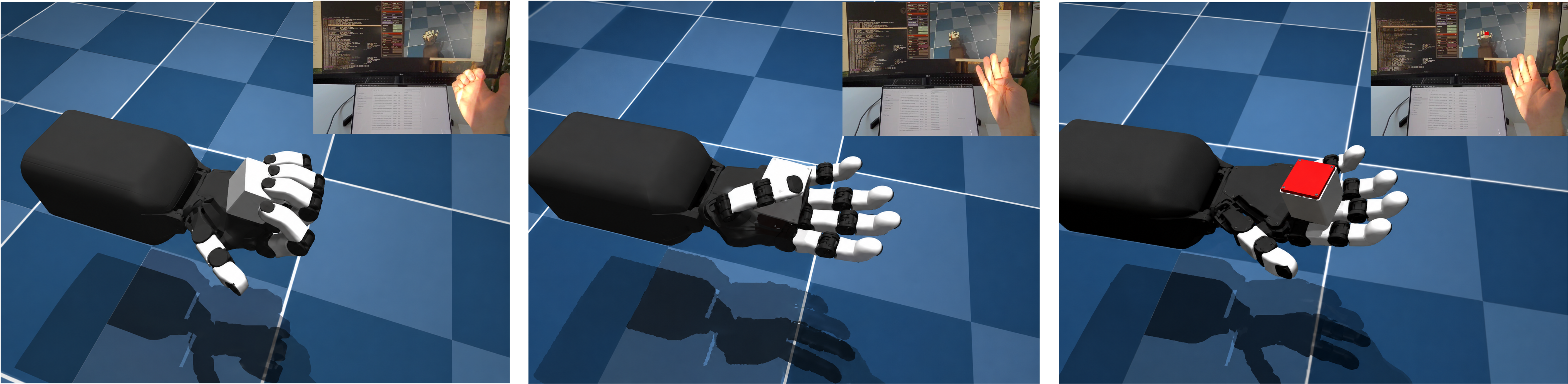}
    \caption{Example of teleoperation for an in-hand orientation task using a Meta Quest 3 headset wired to \orcasim via \orcateleop.}
    \label{fig:metaquest}
\end{figure}

Lastly, we validate the complete workflow for which the \orca~stack is designed on an in-hand reorientation task: the operator teleoperates the (simulated) hand using a Meta Quest device with hand tracking through \orcateleop, recording a total of 10 expert demonstrations (Figure~\ref{fig:metaquest}) that are stored directly in the \texttt{LeRobotDataset} format.
Then, we train a compact behavioral cloning policy based on ACT~\citep{zhaoLearningFineGrainedBimanual2023a} (50M parameters) on this small demonstration data using LeRobot's API only, and with default configuration, without any changes to the learning code.
Observations to the policy include the hand's proprioceptive state, together with wrist and topdown camera views (Figure~\ref{fig:wrist-and-top}).
When evaluated in the same environment, the learned policy succeeds in 9/10 test rollouts, showing that demonstrations collected through the stack are of sufficient quality to learn a closed-loop policy working exclusively from visuo-motor inputs.
Exercising every component of our \orca~software stack (hand interfacing, simulation, teleoperation, retargeting, dataset recording, policy learning and evaluation), this experiment confirms the interoperability of our packages as one complete stack for robot learning research.

%% file: sections/05_conclusions.tex
We presented \orca, a fully open-source stack for dexterous-manipulation research.
\orca~builds on an affordable, 3D-printable, tendon-driven hand with an integrated software suite for control, simulation, full-manipulator integration, retargeting, and teleoperation through consumer-grade devices.
Our stack is natively integrated with \lerobot, bringing dexterous hands onto the same data, training, and evaluation pipeline used for grippers and other affordable robot arms.
The entire stack---from low-level interfaces, to data collection, policy training, and evaluation---is released under the MIT license, and it jointly exercises teleoperation, retargeting, dataset recording, and policy learning within one unified, fully open pipeline.
By reducing the cost of adopting a dexterous hand to that of any other \lerobot-supported robot, our \orca~stack aims at lowering the barrier to entry to dexterous hand, further democratizing dexterity research.
By fully releasing our software stack under a permissive license, we aim to establish \orca~as a shared, affordable, and extensible platform for dexterity research.

\paragraph{Limitations.}
While confident in the potential impact and usefulness of \orca~to lower to barrier to entry to dexterity for the broader research community, we still identify concrete limitations in our contributions.
First, the accesibility of the stack may be hindered by its packaging, as we intentionally distribute the software stack as four independent subpackages rather than a monolith.
Although this choice improves separation of concerns and enables independent feedback channels from the open-source community, it also risks hindering wider accessibility, and increases friction for users. 
In turn, we plan to revisit this design once sufficient feedback allows the interfaces to stabilize.
Second, faithful simulation of tendon-driven, contact-rich interactions remains an open problem: tendon coupling and soft-finger contact are rather difficult to model exactly, and our simulation environments provide only a coarse approximation of the physical hand's interaction with its environment.
Third, our teleoperation pipeline could still be augmented with more, and more diverse publishers.
Lastly, throughout we purposely treat simulation as a first-class domain to enable contributions even from resource-constrained settings (e.g., teleoperating a simulated hand with a simple webcam through MediaPipe), and argue that a simulation-first evaluation pipeline improves accessibility by relaxing hardware requirements.
At the same time, however, the accessibility of the underlying hardware hand makes it so that codesigning our stack for both real and simulated hands might turn out to be suboptimal.
Thus, extending this contribution to (1) a carefully designed monolith package for dexterity, (2) a rich set of simulation tools for tendon-driven hands and contact-rich tasks, together with (3) increased focus on real-world use-cases remain open avenues.

%% file: sections/06_appendix.tex

This appendix complements the main text with a code-first tour of the \orca~software stack: how to install the stack, drive a hand, simulate it, scale to full platforms, and close the loop from teleoperation to policy learning---and answer each with a minimal, executable example.
All snippets run against the publicly released packages.

\subsection{Packages and installation}
\label{app:packages}

\begin{table}[h]
    \centering
    \small
    \begin{tabular}{lll}
        \toprule
        Package & Scope & Distribution \\
        \midrule
        \orcacore & Hand interface, calibration, typed joint-space control & PyPI \& source \\
        \orcasim & MuJoCo hands and task environments (Gymnasium API) & PyPI \& source \\
        \orcaarm~ & URDF/MJCF descriptions of full manipulation platforms & PyPI \& source \\
        \orcateleop & Teleoperation, retargeting, \lerobot~integration & source \\
        \bottomrule
    \end{tabular}
    \caption{The four packages of the \orca~stack and how to obtain them.}
    \label{tab:packages}
\end{table}

\subsection{\orcacore: driving a hand}
\label{app:orcacore}

A physical hand is driven through a single class, with the serial port, motor family, and baudrate auto-detected at connection time (Code example~\ref{alg:app-core}).
Joint commands and readings travel as \texttt{OrcaJointPositions}, an immutable, name-indexed payload asserted at the interface boundary.
Maintenance routines are one-liners on the same object (\texttt{hand.calibrate()}, \texttt{hand.tension()}), and \texttt{MockOrcaHand} exposes the identical API without hardware, so the full control path can be exercised before a hand is ever connected (useful for testing).
For hands mounting sensors such as tactile sensors or joint encoders, the \texttt{OrcaHandTouch} variant adds tactile access (e.g., \texttt{get\_tactile\_forces()}) on top of the same interface.

\begin{algorithm}[H]
\caption{Hand control with \orcacore.}
\label{alg:app-core}
\rule{\linewidth}{1pt}
\begin{lstlisting}
from orca_core import OrcaHand, OrcaJointPositions

hand = OrcaHand()   # bundled hand model; port auto-detected
hand.connect()
hand.init_joints()  # enables torque, calibrates if needed

pose = OrcaJointPositions.from_dict({"thumb_mcp": 0.5, "index_mcp": 0.3})
hand.set_joint_positions(pose, num_steps=10, step_size=0.02)

state = hand.get_joint_position()          # typed, frozen payload
temps = hand.get_motor_temp(as_dict=True)  # per-motor diagnostics
hand.disconnect()

# no hardware yet? identical API, in-memory backend:
from orca_core.hardware_hand import MockOrcaHand
hand = MockOrcaHand()
\end{lstlisting}
\rule{\linewidth}{1pt}
\end{algorithm}

\subsection{\orcasim: simulated counterparts}
\label{app:orcasim}

Simulation is exposed at two levels.
At the task level, environments follow the Gymnasium API and run headlessly by default (Code example~\ref{alg:app-sim}); the in-hand reorientation task presented in the main text exposes a 17-d joint-space action and a 51-d observation, with environments reachable through the Gymnasium registry.
At the hand level, \texttt{SimOrcaHand} subclasses the same \orcacore~base interface as the physical hand, so joint-space code is interchangeable across the real and simulated device (Code example~\ref{alg:app-shared}), substantiating our claim behind the "one control abstraction".

\begin{algorithm}[H]
\caption{Task environments in \orcasim.}
\label{alg:app-sim}
\rule{\linewidth}{1pt}
\begin{lstlisting}
import gymnasium as gym
from orca_sim import OrcaHandRightCubeOrientation, register_envs

env = OrcaHandRightCubeOrientation(render_mode="rgb_array")
obs, info = env.reset(seed=0)         # 51-d observation
for _ in range(200):
    action = env.action_space.sample()  # 17-d joint targets
    obs, reward, terminated, truncated, info = env.step(action)
    if terminated or truncated:
        obs, info = env.reset()
frame = env.render()  # RGB array, headless-friendly - can also render human
env.close()

register_envs()  # alternatively, via the Gymnasium registry
env = gym.make("OrcaHandRight-v2")
\end{lstlisting}
\rule{\linewidth}{1pt}
\end{algorithm}

\begin{algorithm}[H]
\caption{One joint-space interface for real and simulated hands.}
\label{alg:app-shared}
\rule{\linewidth}{1pt}
\begin{lstlisting}
from orca_core import OrcaJointPositions
from orca_sim import SimOrcaHand

hand = SimOrcaHand()  # subclasses the orca_core hand interface
hand.reset()

pose = OrcaJointPositions.from_dict({"index_mcp": 20.0})
hand.set_joint_positions(pose)
state = hand.get_joint_position()
\end{lstlisting}
\rule{\linewidth}{1pt}
\end{algorithm}

\subsection{\orcaarm: full manipulation platforms}
\label{app:orcaarm}

\orcaarm~ships self-contained URDF and MJCF descriptions of complete platforms---bimanual OpenArm with two hands, and single or bimanual Franka Emika Panda setups---exposed as importable path constants with all mesh references resolved (Code example~\ref{alg:app-arm}).
The same files back the \orcasim~task environments (e.g., \texttt{OrcaPandaCubeStacking}) and load unchanged in any URDF-compatible toolchain.

\begin{algorithm}[H]
\caption{Loading full-platform descriptions from \orcaarm.}
\label{alg:app-arm}
\rule{\linewidth}{1pt}
\begin{lstlisting}
import mujoco
import orca_arm

# single-arm Panda + right hand: 24 actuators (7-DoF arm + 17-DoF hand)
model = mujoco.MjModel.from_xml_path(orca_arm.ORCAPANDA_MJCF_PATH)
data = mujoco.MjData(model)

orca_arm.URDF_PATH                   # bimanual OpenArm platform
orca_arm.BIMANUAL_ORCAPANDA_MJCF_PATH  # two Pandas, two hands
\end{lstlisting}
\rule{\linewidth}{1pt}
\end{algorithm}

\subsection{\orcateleop: from human motion to robot commands}
\label{app:orcateleop}

The retargeter described in Section~3 is usable as a standalone library: it consumes 21 hand landmarks in the MANO convention---from any source---and produces joint commands for any \orcacore-compatible hand (Code example~\ref{alg:app-retarget}).
The first frames are used for online scale calibration, during which \texttt{retarget} returns \texttt{None}.

\begin{algorithm}[H]
\caption{Standalone retargeting of hand landmarks.}
\label{alg:app-retarget}
\rule{\linewidth}{1pt}
\begin{lstlisting}
from orca_teleop.retargeting import Retargeter, TargetPose

retargeter = Retargeter.from_paths(backend="adaptive_analytical")
landmarks = ...  # (21, 3) MANO-convention hand landmarks
joints = retargeter.retarget(TargetPose(joint_positions=landmarks))
# -> OrcaJointPositions, valid for any orca_core-compatible hand
\end{lstlisting}
\rule{\linewidth}{1pt}
\end{algorithm}

The full pipeline composes a landmark publisher, the retargeter, and a robot sink; publishers stream landmarks over gRPC, so operator and robot can sit on one machine or two, and the sink selects the target embodiment---real or simulated---without touching the rest of the pipeline (Code example~\ref{alg:app-teleop}).

\begin{algorithm}[H]
\caption{Teleoperation pipelines in \orcateleop.}
\label{alg:app-teleop}
\rule{\linewidth}{1pt}
\begin{lstlisting}
from orca_teleop.pipeline import run, run_local, OrcaHandSink
from orca_teleop.sim import OrcaHandSimSink

# single machine: webcam (MediaPipe) -> retargeter -> simulated hand
run_local(sink=OrcaHandSimSink(env_name="right"), show_video=True)

# distributed: on the operator machine, stream landmarks over gRPC...
from orca_teleop.ingress.mediapipe.publisher import MediaPipePublisher
MediaPipePublisher(server_address="<robot-ip>:50051").run()
# ...while on the robot machine the same pipeline drives the real hand
run(sink=OrcaHandSink(model_path=None), port=50051)
\end{lstlisting}
\rule{\linewidth}{1pt}
\end{algorithm}

Demonstrations are recorded directly in the \texttt{LeRobotDataset} format from either backend (Code example~\ref{alg:app-record}), and trained \lerobot~policies can be rolled out closed-loop through the same sink abstraction used for teleoperation (Code example~\ref{alg:app-policy}).
The adapter is policy-agnostic: the checkpoint declares its policy type, which is resolved through \lerobot's policy factory, so the same rollout code serves ACT, Diffusion Policy, or \( \pi_0 \) checkpoints alike.
Recording and rollout share one observation contract, in which the textual task description is always present: state-only policies simply ignore it, while language-conditioned policies consume it without any change to the calling code.

\begin{algorithm}[H]
\caption{Recording demonstrations in the \texttt{LeRobotDataset} format.}
\label{alg:app-record}
\rule{\linewidth}{1pt}
\begin{lstlisting}[language=bash]
python scripts/record_dataset.py \
    --repo-id $HF_USER/orca-cube --task "rotate the cube" \
    --num-episodes 10 --backend sim \
    --camera top:0 --push-to-hub
\end{lstlisting}
\rule{\linewidth}{1pt}
\end{algorithm}

\begin{algorithm}[H]
\caption{Closed-loop rollout of a \lerobot~policy.}
\label{alg:app-policy}
\rule{\linewidth}{1pt}
\begin{lstlisting}
from orca_teleop.policies import LeRobotPolicyAdapter
from orca_teleop.sim import OrcaHandSimSink

# policy class (ACT, Diffusion Policy, pi0, ...) resolved from checkpoint
policy = LeRobotPolicyAdapter.from_pretrained(
    "outputs/train/checkpoints/last/pretrained_model",
    dataset_repo_id="<user>/orca-cube",  # training data (norm. stats)
)
sink = OrcaHandSimSink(env_name="right")
sink.connect()
for _ in range(500):
    sink.step_policy(policy, task="rotate the cube")
\end{lstlisting}
\rule{\linewidth}{1pt}
\end{algorithm}

%% file: references.bib
@article{akkaya2019solving,
  title = {Solving Rubik's Cube with a Robot Hand},
  author = {Akkaya, Ilge and Andrychowicz, Marcin and Chociej, Maciek and Litwin, Mateusz and McGrew, Bob and Petron, Arthur and Paino, Alex and Plappert, Matthias and Powell, Glenn and Ribas, Raphael and others},
  year = 2019,
  journal = {arXiv preprint arXiv:1910.07113},
  eprint = {1910.07113},
  archiveprefix = {arXiv}
}

@misc{antonovaReinforcementLearningPivoting2017a,
  title = {Reinforcement {{Learning}} for {{Pivoting Task}}},
  author = {Antonova, Rika and Cruciani, Silvia and Smith, Christian and Kragic, Danica},
  year = 2017,
  month = mar,
  number = {arXiv:1703.00472},
  eprint = {1703.00472},
  primaryclass = {cs.RO},
  publisher = {arXiv},
  doi = {10.48550/arXiv.1703.00472},
  urldate = {2026-05-13},
  archiveprefix = {arXiv}
}

@misc{black$p_0$VisionLanguageActionFlow2026,
  title = {\${$\pi\_$}0\$: {{A Vision-Language-Action Flow Model}} for {{General Robot Control}}},
  shorttitle = {\${$\pi\_$}0\$},
  author = {Black, Kevin and Brown, Noah and Driess, Danny and Esmail, Adnan and Equi, Michael and Finn, Chelsea and Fusai, Niccolo and Groom, Lachy and Hausman, Karol and Ichter, Brian and Jakubczak, Szymon and Jones, Tim and Ke, Liyiming and Levine, Sergey and {Li-Bell}, Adrian and Mothukuri, Mohith and Nair, Suraj and Pertsch, Karl and Shi, Lucy Xiaoyang and Tanner, James and Vuong, Quan and Walling, Anna and Wang, Haohuan and Zhilinsky, Ury},
  year = 2026,
  month = jan,
  number = {arXiv:2410.24164},
  eprint = {2410.24164},
  primaryclass = {cs.LG},
  publisher = {arXiv},
  doi = {10.48550/arXiv.2410.24164},
  urldate = {2026-05-28},
  archiveprefix = {arXiv}
}

@misc{brohanRT1RoboticsTransformer2023a,
  title = {{{RT-1}}: {{Robotics Transformer}} for {{Real-World Control}} at {{Scale}}},
  shorttitle = {{{RT-1}}},
  author = {Brohan, Anthony and Brown, Noah and Carbajal, Justice and Chebotar, Yevgen and Dabis, Joseph and Finn, Chelsea and Gopalakrishnan, Keerthana and Hausman, Karol and Herzog, Alex and Hsu, Jasmine and Ibarz, Julian and Ichter, Brian and Irpan, Alex and Jackson, Tomas and Jesmonth, Sally and Joshi, Nikhil J. and Julian, Ryan and Kalashnikov, Dmitry and Kuang, Yuheng and Leal, Isabel and Lee, Kuang-Huei and Levine, Sergey and Lu, Yao and Malla, Utsav and Manjunath, Deeksha and Mordatch, Igor and Nachum, Ofir and Parada, Carolina and Peralta, Jodilyn and Perez, Emily and Pertsch, Karl and Quiambao, Jornell and Rao, Kanishka and Ryoo, Michael and Salazar, Grecia and Sanketi, Pannag and Sayed, Kevin and Singh, Jaspiar and Sontakke, Sumedh and Stone, Austin and Tan, Clayton and Tran, Huong and Vanhoucke, Vincent and Vega, Steve and Vuong, Quan and Xia, Fei and Xiao, Ted and Xu, Peng and Xu, Sichun and Yu, Tianhe and Zitkovich, Brianna},
  year = 2023,
  month = aug,
  number = {arXiv:2212.06817},
  eprint = {2212.06817},
  primaryclass = {cs.RO},
  publisher = {arXiv},
  doi = {10.48550/arXiv.2212.06817},
  urldate = {2026-05-12},
  archiveprefix = {arXiv}
}

@misc{brohanRT2VisionLanguageActionModels2023a,
  title = {{{RT-2}}: {{Vision-Language-Action Models Transfer Web Knowledge}} to {{Robotic Control}}},
  shorttitle = {{{RT-2}}},
  author = {Brohan, Anthony and Brown, Noah and Carbajal, Justice and Chebotar, Yevgen and Chen, Xi and Choromanski, Krzysztof and Ding, Tianli and Driess, Danny and Dubey, Avinava and Finn, Chelsea and Florence, Pete and Fu, Chuyuan and Arenas, Montse Gonzalez and Gopalakrishnan, Keerthana and Han, Kehang and Hausman, Karol and Herzog, Alexander and Hsu, Jasmine and Ichter, Brian and Irpan, Alex and Joshi, Nikhil and Julian, Ryan and Kalashnikov, Dmitry and Kuang, Yuheng and Leal, Isabel and Lee, Lisa and Lee, Tsang-Wei Edward and Levine, Sergey and Lu, Yao and Michalewski, Henryk and Mordatch, Igor and Pertsch, Karl and Rao, Kanishka and Reymann, Krista and Ryoo, Michael and Salazar, Grecia and Sanketi, Pannag and Sermanet, Pierre and Singh, Jaspiar and Singh, Anikait and Soricut, Radu and Tran, Huong and Vanhoucke, Vincent and Vuong, Quan and Wahid, Ayzaan and Welker, Stefan and Wohlhart, Paul and Wu, Jialin and Xia, Fei and Xiao, Ted and Xu, Peng and Xu, Sichun and Yu, Tianhe and Zitkovich, Brianna},
  year = 2023,
  month = jul,
  number = {arXiv:2307.15818},
  eprint = {2307.15818},
  primaryclass = {cs.RO},
  publisher = {arXiv},
  doi = {10.48550/arXiv.2307.15818},
  urldate = {2026-05-12},
  archiveprefix = {arXiv}
}

@misc{cadeneLeRobotOpenSourceLibrary2026,
  title = {{{LeRobot}}: {{An Open-Source Library}} for {{End-to-End Robot Learning}}},
  shorttitle = {{{LeRobot}}},
  author = {Cadene, Remi and Aliberts, Simon and Capuano, Francesco and Aractingi, Michel and Zouitine, Adil and Kooijmans, Pepijn and Choghari, Jade and Russi, Martino and Pascal, Caroline and Palma, Steven and Shukor, Mustafa and Moss, Jess and Soare, Alexander and Aubakirova, Dana and Lhoest, Quentin and Gallou{\'e}dec, Quentin and Wolf, Thomas},
  year = 2026,
  month = feb,
  number = {arXiv:2602.22818},
  eprint = {2602.22818},
  primaryclass = {cs.RO},
  publisher = {arXiv},
  doi = {10.48550/arXiv.2602.22818},
  urldate = {2026-05-12},
  archiveprefix = {arXiv}
}

@article{capuano2025robot,
  title = {Robot Learning: {{A}} Tutorial},
  author = {Capuano, Francesco and Pascal, Caroline and Zouitine, Adil and Wolf, Thomas and Aractingi, Michel},
  year = 2025,
  journal = {arXiv preprint arXiv:2510.12403},
  eprint = {2510.12403},
  archiveprefix = {arXiv}
}

@misc{chenSystemGeneralInHand2021,
  title = {A {{System}} for {{General In-Hand Object Re-Orientation}}},
  author = {Chen, Tao and Xu, Jie and Agrawal, Pulkit},
  year = 2021,
  month = nov,
  number = {arXiv:2111.03043},
  eprint = {2111.03043},
  primaryclass = {cs.RO},
  publisher = {arXiv},
  doi = {10.48550/arXiv.2111.03043},
  urldate = {2026-05-13},
  archiveprefix = {arXiv}
}

@misc{chiDiffusionPolicyVisuomotor2024a,
  title = {Diffusion {{Policy}}: {{Visuomotor Policy Learning}} via {{Action Diffusion}}},
  shorttitle = {Diffusion {{Policy}}},
  author = {Chi, Cheng and Xu, Zhenjia and Feng, Siyuan and Cousineau, Eric and Du, Yilun and Burchfiel, Benjamin and Tedrake, Russ and Song, Shuran},
  year = 2024,
  month = mar,
  number = {arXiv:2303.04137},
  eprint = {2303.04137},
  primaryclass = {cs.RO},
  publisher = {arXiv},
  doi = {10.48550/arXiv.2303.04137},
  urldate = {2026-05-12},
  archiveprefix = {arXiv}
}

@misc{christophORCAOpenSourceReliable2025,
  title = {{{ORCA}}: {{An Open-Source}}, {{Reliable}}, {{Cost-Effective}}, {{Anthropomorphic Robotic Hand}} for {{Uninterrupted Dexterous Task Learning}}},
  shorttitle = {{{ORCA}}},
  author = {Christoph, Clemens C. and Eberlein, Maximilian and Katsimalis, Filippos and Roberti, Arturo and Sympetheros, Aristotelis and Vogt, Michel R. and Liconti, Davide and Yang, Chenyu and Cangan, Barnabas Gavin and Hinchet, Ronan J. and Katzschmann, Robert K.},
  year = 2025,
  month = sep,
  number = {arXiv:2504.04259},
  eprint = {2504.04259},
  primaryclass = {cs.RO},
  publisher = {arXiv},
  doi = {10.48550/arXiv.2504.04259},
  urldate = {2026-05-14},
  archiveprefix = {arXiv}
}

@misc{collaborationOpenXEmbodimentRobotic2025,
  title = {Open {{X-Embodiment}}: {{Robotic Learning Datasets}} and {{RT-X Models}}},
  shorttitle = {Open {{X-Embodiment}}},
  author = {Collaboration, Open X.-Embodiment and O'Neill, Abby and Rehman, Abdul and Gupta, Abhinav and Maddukuri, Abhiram and Gupta, Abhishek and Padalkar, Abhishek and Lee, Abraham and Pooley, Acorn and Gupta, Agrim and Mandlekar, Ajay and Jain, Ajinkya and Tung, Albert and Bewley, Alex and Herzog, Alex and Irpan, Alex and Khazatsky, Alexander and Rai, Anant and Gupta, Anchit and Wang, Andrew and Kolobov, Andrey and Singh, Anikait and Garg, Animesh and Kembhavi, Aniruddha and Xie, Annie and Brohan, Anthony and Raffin, Antonin and Sharma, Archit and Yavary, Arefeh and Jain, Arhan and Balakrishna, Ashwin and Wahid, Ayzaan and {Burgess-Limerick}, Ben and Kim, Beomjoon and Sch{\"o}lkopf, Bernhard and Wulfe, Blake and Ichter, Brian and Lu, Cewu and Xu, Charles and Le, Charlotte and Finn, Chelsea and Wang, Chen and Xu, Chenfeng and Chi, Cheng and Huang, Chenguang and Chan, Christine and Agia, Christopher and Pan, Chuer and Fu, Chuyuan and Devin, Coline and Xu, Danfei and Morton, Daniel and Driess, Danny and Chen, Daphne and Pathak, Deepak and Shah, Dhruv and B{\"u}chler, Dieter and Jayaraman, Dinesh and Kalashnikov, Dmitry and Sadigh, Dorsa and Johns, Edward and Foster, Ethan and Liu, Fangchen and Ceola, Federico and Xia, Fei and Zhao, Feiyu and Frujeri, Felipe Vieira and Stulp, Freek and Zhou, Gaoyue and Sukhatme, Gaurav S. and Salhotra, Gautam and Yan, Ge and Feng, Gilbert and Schiavi, Giulio and Berseth, Glen and Kahn, Gregory and Yang, Guangwen and Wang, Guanzhi and Su, Hao and Fang, Hao-Shu and Shi, Haochen and Bao, Henghui and Amor, Heni Ben and Christensen, Henrik I. and Furuta, Hiroki and Bharadhwaj, Homanga and Walke, Homer and Fang, Hongjie and Ha, Huy and Mordatch, Igor and Radosavovic, Ilija and Leal, Isabel and Liang, Jacky and {Abou-Chakra}, Jad and Kim, Jaehyung and Drake, Jaimyn and Peters, Jan and Schneider, Jan and Hsu, Jasmine and Vakil, Jay and Bohg, Jeannette and Bingham, Jeffrey and Wu, Jeffrey and Gao, Jensen and Hu, Jiaheng and Wu, Jiajun and Wu, Jialin and Sun, Jiankai and Luo, Jianlan and Gu, Jiayuan and Tan, Jie and Oh, Jihoon and Wu, Jimmy and Lu, Jingpei and Yang, Jingyun and Malik, Jitendra and Silv{\'e}rio, Jo{\~a}o and Hejna, Joey and Booher, Jonathan and Tompson, Jonathan and Yang, Jonathan and Salvador, Jordi and Lim, Joseph J. and Han, Junhyek and Wang, Kaiyuan and Rao, Kanishka and Pertsch, Karl and Hausman, Karol and Go, Keegan and Gopalakrishnan, Keerthana and Goldberg, Ken and Byrne, Kendra and Oslund, Kenneth and Kawaharazuka, Kento and Black, Kevin and Lin, Kevin and Zhang, Kevin and Ehsani, Kiana and Lekkala, Kiran and Ellis, Kirsty and Rana, Krishan and Srinivasan, Krishnan and Fang, Kuan and Singh, Kunal Pratap and Zeng, Kuo-Hao and Hatch, Kyle and Hsu, Kyle and Itti, Laurent and Chen, Lawrence Yunliang and Pinto, Lerrel and {Fei-Fei}, Li and Tan, Liam and Fan, Linxi "Jim" and Ott, Lionel and Lee, Lisa and Weihs, Luca and Chen, Magnum and Lepert, Marion and Memmel, Marius and Tomizuka, Masayoshi and Itkina, Masha and Castro, Mateo Guaman and Spero, Max and Du, Maximilian and Ahn, Michael and Yip, Michael C. and Zhang, Mingtong and Ding, Mingyu and Heo, Minho and Srirama, Mohan Kumar and Sharma, Mohit and Kim, Moo Jin and Irshad, Muhammad Zubair and Kanazawa, Naoaki and Hansen, Nicklas and Heess, Nicolas and Joshi, Nikhil J. and Suenderhauf, Niko and Liu, Ning and Palo, Norman Di and Shafiullah, Nur Muhammad Mahi and Mees, Oier and Kroemer, Oliver and Bastani, Osbert and Sanketi, Pannag R. and Miller, Patrick "Tree" and Yin, Patrick and Wohlhart, Paul and Xu, Peng and Fagan, Peter David and Mitrano, Peter and Sermanet, Pierre and Abbeel, Pieter and Sundaresan, Priya and Chen, Qiuyu and Vuong, Quan and Rafailov, Rafael and Tian, Ran and Doshi, Ria and {Mart{\'i}n-Mart{\'i}n}, Roberto and Baijal, Rohan and Scalise, Rosario and Hendrix, Rose and Lin, Roy and Qian, Runjia and Zhang, Ruohan and Mendonca, Russell and Shah, Rutav and Hoque, Ryan and Julian, Ryan and Bustamante, Samuel and Kirmani, Sean and Levine, Sergey and Lin, Shan and Moore, Sherry and Bahl, Shikhar and Dass, Shivin and Sonawani, Shubham and Tulsiani, Shubham and Song, Shuran and Xu, Sichun and Haldar, Siddhant and Karamcheti, Siddharth and Adebola, Simeon and Guist, Simon and Nasiriany, Soroush and Schaal, Stefan and Welker, Stefan and Tian, Stephen and Ramamoorthy, Subramanian and Dasari, Sudeep and Belkhale, Suneel and Park, Sungjae and Nair, Suraj and Mirchandani, Suvir and Osa, Takayuki and Gupta, Tanmay and Harada, Tatsuya and Matsushima, Tatsuya and Xiao, Ted and Kollar, Thomas and Yu, Tianhe and Ding, Tianli and Davchev, Todor and Zhao, Tony Z. and Armstrong, Travis and Darrell, Trevor and Chung, Trinity and Jain, Vidhi and Kumar, Vikash and Vanhoucke, Vincent and Guizilini, Vitor and Zhan, Wei and Zhou, Wenxuan and Burgard, Wolfram and Chen, Xi and Chen, Xiangyu and Wang, Xiaolong and Zhu, Xinghao and Geng, Xinyang and Liu, Xiyuan and Liangwei, Xu and Li, Xuanlin and Pang, Yansong and Lu, Yao and Ma, Yecheng Jason and Kim, Yejin and Chebotar, Yevgen and Zhou, Yifan and Zhu, Yifeng and Wu, Yilin and Xu, Ying and Wang, Yixuan and Bisk, Yonatan and Dou, Yongqiang and Cho, Yoonyoung and Lee, Youngwoon and Cui, Yuchen and Cao, Yue and Wu, Yueh-Hua and Tang, Yujin and Zhu, Yuke and Zhang, Yunchu and Jiang, Yunfan and Li, Yunshuang and Li, Yunzhu and Iwasawa, Yusuke and Matsuo, Yutaka and Ma, Zehan and Xu, Zhuo and Cui, Zichen Jeff and Zhang, Zichen and Fu, Zipeng and Lin, Zipeng},
  year = 2025,
  month = may,
  number = {arXiv:2310.08864},
  eprint = {2310.08864},
  primaryclass = {cs.RO},
  publisher = {arXiv},
  doi = {10.48550/arXiv.2310.08864},
  urldate = {2026-05-12},
  archiveprefix = {arXiv}
}

@misc{florenceImplicitBehavioralCloning2021,
  title = {Implicit {{Behavioral Cloning}}},
  author = {Florence, Pete and Lynch, Corey and Zeng, Andy and Ramirez, Oscar and Wahid, Ayzaan and Downs, Laura and Wong, Adrian and Lee, Johnny and Mordatch, Igor and Tompson, Jonathan},
  year = 2021,
  month = sep,
  number = {arXiv:2109.00137},
  eprint = {2109.00137},
  primaryclass = {cs.RO},
  publisher = {arXiv},
  doi = {10.48550/arXiv.2109.00137},
  urldate = {2026-05-12},
  archiveprefix = {arXiv}
}

@misc{frankaemika_panda,
  title = {Franka Emika Panda Robot},
  author = {{Franka Emika GmbH}},
  year = 2017
}

@misc{kareerEgoMimicScalingImitation2024,
  title = {{{EgoMimic}}: {{Scaling Imitation Learning}} via {{Egocentric Video}}},
  shorttitle = {{{EgoMimic}}},
  author = {Kareer, Simar and Patel, Dhruv and Punamiya, Ryan and Mathur, Pranay and Cheng, Shuo and Wang, Chen and Hoffman, Judy and Xu, Danfei},
  year = 2024,
  month = oct,
  number = {arXiv:2410.24221},
  eprint = {2410.24221},
  primaryclass = {cs.RO},
  publisher = {arXiv},
  doi = {10.48550/arXiv.2410.24221},
  urldate = {2026-05-14},
  archiveprefix = {arXiv},
  langid = {english}
}

@misc{kediaSimToolRealObjectCentricPolicy2026a,
  title = {{{SimToolReal}}: {{An Object-Centric Policy}} for {{Zero-Shot Dexterous Tool Manipulation}}},
  shorttitle = {{{SimToolReal}}},
  author = {Kedia, Kushal and Lum, Tyler Ga Wei and Bohg, Jeannette and Liu, C. Karen},
  year = 2026,
  month = feb,
  number = {arXiv:2602.16863},
  eprint = {2602.16863},
  primaryclass = {cs.RO},
  publisher = {arXiv},
  doi = {10.48550/arXiv.2602.16863},
  urldate = {2026-05-13},
  archiveprefix = {arXiv}
}

@inproceedings{khazatskyDROIDLargeScaleInTheWild2024,
  title = {{{DROID}}: {{A Large-Scale In-The-Wild Robot Manipulation Dataset}}},
  shorttitle = {{{DROID}}},
  booktitle = {Robotics: {{Science}} and {{Systems XX}}},
  author = {Khazatsky, Alexander and Pertsch, Karl and Nair, Suraj and Balakrishna, Ashwin and Dasari, Sudeep and Karamcheti, Siddharth and Nasiriany, Soroush and Srirama, Mohan and Chen, Lawrence and Ellis, Kirsty and Fagan, Peter and Hejna, Joey and Itkina, Masha and Lepert, Marion and Ma, Yecheng and Miller, Patrick and Wu, Jimmy and Belkhale, Suneel and Dass, Shivin and Ha, Huy and Jain, Arhan and Lee, Abraham and Lee, Youngwoon and Memmel, Marius and Park, Sungjae and Radosavovic, Ilija and Wang, Kaiyuan and Zhan, Albert and Black, Kevin and Chi, Cheng and Hatch, Kyle and Lin, Shan and Lu, Jingpei and Mercat, Jean and Rehman, Abdul and Sanketi, Pannag and Sharma, Archit and Simpson, Cody and Vuong, Quan and Walke, Homer and Wulfe, Blake and Xiao, Ted and Yang, Jonathan and Yavary, Arefeh and Zhao, Tony and Agia, Christopher and Baijal, Rohan and Castro, Mateo and Chen, Daphne and Chen, Qiuyu and Chung, Trinity and Drake, Jaimyn and Foster, Ethan and Gao, Jensen and Herrera, David and Heo, Minho and Hsu, Kyle and Hu, Jiaheng and Jackson, Donovon and Le, Charlotte and Li, Yunshuang and Lin, Roy and Ma, Zehan and Maddukuri, Abhiram and Mirchandani, Suvir and Morton, Daniel and Nguyen, Tony and O'Neill, Abigail and Scalise, Rosario and Seale, Derick and Son, Victor and Tian, Stephen and Tran, Emi and Wang, Andrew and Wu, Yilin and Xie, Annie and Yang, Jingyun and Yin, Patrick and Zhang, Yunchu and Bastani, Osbert and Berseth, Glen and Bohg, Jeannette and Goldberg, Ken and Gupta, Abhinav and Gupta, Abhishek and Jayaraman, Dinesh and Lim, Joseph and Malik, Jitendra and {Mart{\'i}n-Mart{\'i}n}, Roberto and Ramamoorthy, Subramanian and Sadigh, Dorsa and Song, Shuran and Wu, Jiajun and Yip, Michael and Zhu, Yuke and Kollar, Thomas and Levine, Sergey and Finn, Chelsea},
  year = 2024,
  month = jul,
  publisher = {{Robotics: Science and Systems Foundation}},
  doi = {10.15607/RSS.2024.XX.120},
  urldate = {2026-05-13},
  isbn = {979-8-9902848-0-7},
  langid = {english}
}

@misc{kimOpenVLAOpenSourceVisionLanguageAction2024a,
  title = {{{OpenVLA}}: {{An Open-Source Vision-Language-Action Model}}},
  shorttitle = {{{OpenVLA}}},
  author = {Kim, Moo Jin and Pertsch, Karl and Karamcheti, Siddharth and Xiao, Ted and Balakrishna, Ashwin and Nair, Suraj and Rafailov, Rafael and Foster, Ethan and Lam, Grace and Sanketi, Pannag and Vuong, Quan and Kollar, Thomas and Burchfiel, Benjamin and Tedrake, Russ and Sadigh, Dorsa and Levine, Sergey and Liang, Percy and Finn, Chelsea},
  year = 2024,
  month = sep,
  number = {arXiv:2406.09246},
  eprint = {2406.09246},
  primaryclass = {cs.RO},
  publisher = {arXiv},
  doi = {10.48550/arXiv.2406.09246},
  urldate = {2026-05-12},
  archiveprefix = {arXiv}
}

@article{kober2013reinforcement,
  title = {Reinforcement Learning in Robotics: {{A}} Survey},
  author = {Kober, Jens and Bagnell, J Andrew and Peters, Jan},
  year = 2013,
  journal = {The International Journal of Robotics Research},
  volume = {32},
  number = {11},
  pages = {1238--1274},
  publisher = {SAGE Publications Sage UK: London, England}
}

@misc{liuLIBEROBenchmarkingKnowledge2023a,
  title = {{{LIBERO}}: {{Benchmarking Knowledge Transfer}} for {{Lifelong Robot Learning}}},
  shorttitle = {{{LIBERO}}},
  author = {Liu, Bo and Zhu, Yifeng and Gao, Chongkai and Feng, Yihao and Liu, Qiang and Zhu, Yuke and Stone, Peter},
  year = 2023,
  month = oct,
  number = {arXiv:2306.03310},
  eprint = {2306.03310},
  primaryclass = {cs.AI},
  publisher = {arXiv},
  doi = {10.48550/arXiv.2306.03310},
  urldate = {2026-05-12},
  archiveprefix = {arXiv}
}

@article{lugaresiMediaPipeFrameworkBuilding,
  title = {{{MediaPipe}}: {{A Framework}} for {{Building Perception Pipelines}}},
  author = {Lugaresi, Camillo and Tang, Jiuqiang and Nash, Hadon and McClanahan, Chris and Uboweja, Esha and Hays, Michael and Zhang, Fan and Chang, Chuo-Ling and Yong, Ming Guang and Lee, Juhyun and Chang, Wan-Teh and Hua, Wei and Georg, Manfred and Grundmann, Matthias},
  langid = {english}
}

@misc{manus_metagloves_pro,
  title = {Metagloves pro: {{High-precision}}, Low-Latency Data Gloves},
  author = {{MANUS Technology Group}},
  year = 2026
}

@misc{meta_quest3,
  title = {Meta Quest 3},
  author = {{Meta Platforms, Inc.}},
  year = 2023
}

@misc{muManiSkillGeneralizableManipulation2021,
  title = {{{ManiSkill}}: {{Generalizable Manipulation Skill Benchmark}} with {{Large-Scale Demonstrations}}},
  shorttitle = {{{ManiSkill}}},
  author = {Mu, Tongzhou and Ling, Zhan and Xiang, Fanbo and Yang, Derek and Li, Xuanlin and Tao, Stone and Huang, Zhiao and Jia, Zhiwei and Su, Hao},
  year = 2021,
  month = nov,
  number = {arXiv:2107.14483},
  eprint = {2107.14483},
  primaryclass = {cs.LG},
  publisher = {arXiv},
  doi = {10.48550/arXiv.2107.14483},
  urldate = {2026-05-26},
  archiveprefix = {arXiv}
}

@misc{openarm,
  title = {{{OpenArm}}: A Fully Open-Source Humanoid Arm for Physical {{AI}} Research and Deployment in Contact-Rich Environments},
  author = {{Enactic, Inc.}},
  year = 2026
}

@misc{shukorSmolVLAVisionLanguageActionModel2025a,
  title = {{{SmolVLA}}: {{A Vision-Language-Action Model}} for {{Affordable}} and {{Efficient Robotics}}},
  shorttitle = {{{SmolVLA}}},
  author = {Shukor, Mustafa and Aubakirova, Dana and Capuano, Francesco and Kooijmans, Pepijn and Palma, Steven and Zouitine, Adil and Aractingi, Michel and Pascal, Caroline and Russi, Martino and Marafioti, Andres and Alibert, Simon and Cord, Matthieu and Wolf, Thomas and Cadene, Remi},
  year = 2025,
  month = jun,
  number = {arXiv:2506.01844},
  eprint = {2506.01844},
  primaryclass = {cs.LG},
  publisher = {arXiv},
  doi = {10.48550/arXiv.2506.01844},
  urldate = {2026-05-12},
  archiveprefix = {arXiv}
}

@inproceedings{todorovMuJoCoPhysicsEngine2012,
  title = {{{MuJoCo}}: {{A}} Physics Engine for Model-Based Control},
  shorttitle = {{{MuJoCo}}},
  booktitle = {2012 {{IEEE}}/{{RSJ International Conference}} on {{Intelligent Robots}} and {{Systems}}},
  author = {Todorov, Emanuel and Erez, Tom and Tassa, Yuval},
  year = 2012,
  month = oct,
  pages = {5026--5033},
  issn = {2153-0866},
  doi = {10.1109/IROS.2012.6386109},
  urldate = {2026-05-26}
}

@misc{vedderCaseHumanHands,
  type = {Personal},
  title = {The {{Case Against Human Hands}}},
  author = {Vedder, Kyle},
  journal = {The Case Against Human Hands},
  urldate = {2026-05-13},
  langid = {english}
}

@misc{wangDexCapScalablePortable2024,
  title = {{{DexCap}}: {{Scalable}} and {{Portable Mocap Data Collection System}} for {{Dexterous Manipulation}}},
  shorttitle = {{{DexCap}}},
  author = {Wang, Chen and Shi, Haochen and Wang, Weizhuo and Zhang, Ruohan and {Fei-Fei}, Li and Liu, C. Karen},
  year = 2024,
  month = mar,
  journal = {arXiv.org},
  urldate = {2026-05-14},
  howpublished = {https://arxiv.org/abs/2403.07788v2},
  langid = {english}
}

@misc{wuGELLOGeneralLowCost2024a,
  title = {{{GELLO}}: {{A General}}, {{Low-Cost}}, and {{Intuitive Teleoperation Framework}} for {{Robot Manipulators}}},
  shorttitle = {{{GELLO}}},
  author = {Wu, Philipp and Shentu, Yide and Yi, Zhongke and Lin, Xingyu and Abbeel, Pieter},
  year = 2024,
  month = jul,
  number = {arXiv:2309.13037},
  eprint = {2309.13037},
  primaryclass = {cs.RO},
  publisher = {arXiv},
  doi = {10.48550/arXiv.2309.13037},
  urldate = {2026-05-12},
  archiveprefix = {arXiv}
}

@misc{wuji2026retargeting,
  title = {{{WujiHand}} Retargeting},
  author = {He, Guanqi and Zhang, Wentao},
  year = 2026
}

@misc{yangEgoVLALearningVisionLanguageAction2025,
  title = {{{EgoVLA}}: {{Learning Vision-Language-Action Models}} from {{Egocentric Human Videos}}},
  shorttitle = {{{EgoVLA}}},
  author = {Yang, Ruihan and Yu, Qinxi and Wu, Yecheng and Yan, Rui and Li, Borui and Cheng, An-Chieh and Zou, Xueyan and Fang, Yunhao and Cheng, Xuxin and Qiu, Ri-Zhao and Yin, Hongxu and Liu, Sifei and Han, Song and Lu, Yao and Wang, Xiaolong},
  year = 2025,
  publisher = {arXiv},
  doi = {10.48550/ARXIV.2507.12440},
  urldate = {2026-05-14},
  copyright = {Creative Commons Attribution 4.0 International},
  langid = {english}
}

@misc{zhaoLearningFineGrainedBimanual2023a,
  title = {Learning {{Fine-Grained Bimanual Manipulation}} with {{Low-Cost Hardware}}},
  author = {Zhao, Tony Z. and Kumar, Vikash and Levine, Sergey and Finn, Chelsea},
  year = 2023,
  month = apr,
  number = {arXiv:2304.13705},
  eprint = {2304.13705},
  primaryclass = {cs.RO},
  publisher = {arXiv},
  doi = {10.48550/arXiv.2304.13705},
  urldate = {2026-05-12},
  archiveprefix = {arXiv}
}

@article{zhengEgoscaleScalingDexterous2026,
  title = {Egoscale: {{Scaling}} Dexterous Manipulation with Diverse Egocentric Human Data},
  shorttitle = {Egoscale},
  author = {Zheng, Ruijie and Niu, Dantong and Xie, Yuqi and Wang, Jing and Xu, Mengda and Jiang, Yunfan and Casta{\~n}eda, Fernando and Hu, Fengyuan and Tan, You Liang and Fu, Letian},
  year = 2026,
  journal = {arXiv preprint arXiv:2602.16710},
  eprint = {2602.16710},
  urldate = {2026-05-14},
  archiveprefix = {arXiv}
}
